\documentclass[conference]{IEEEtran}
\usepackage{cite}
\usepackage{amsmath,amssymb,amsfonts}
\usepackage{algorithmic}
\usepackage{graphicx}
\usepackage{textcomp}
\usepackage{xcolor}
\usepackage{hyperref}
\usepackage{url}
\usepackage{times}
\usepackage{array}
\usepackage{paralist}
\usepackage{float}
\usepackage[utf8]{inputenc}
\usepackage{threeparttable}
\usepackage{booktabs}
\usepackage[T1]{fontenc}
\usepackage{tabularx}
\usepackage{ragged2e}
\usepackage{adjustbox}
\usepackage{rotating}
\usepackage{subcaption}
\usepackage{colortbl} 
\usepackage{pgf}      

\definecolor{lightred}{RGB}{255, 204, 204}
\definecolor{lightblue}{RGB}{204, 204, 255}
\definecolor{white}{RGB}{255, 255, 255}

\newcommand{\colcellbuffer}{\rule{-0.33em}{2ex}}

\newcommand{\colcell}[1]{%
    \ifnum #1>50
        \pgfmathsetmacro{\redComponent}{2*(#1-50)}
        \edef\clrmacro{\noexpand\cellcolor{lightred!\redComponent!white}}\clrmacro{\colcellbuffer\textbf{#1}\colcellbuffer}
    \else
        \pgfmathsetmacro{\blueComponent}{2*(50-#1)}
        \edef\clrmacro{\noexpand\cellcolor{lightblue!\blueComponent!white}}\clrmacro{\colcellbuffer#1\colcellbuffer}
    \fi
}

\newcommand{\colcellnobold}[1]{%
    \ifnum #1>50
        \pgfmathsetmacro{\redComponent}{2*(#1-50)}
        \edef\clrmacro{\noexpand\cellcolor{lightred!\redComponent!white}}\clrmacro{\colcellbuffer#1\colcellbuffer}
    \else
        \pgfmathsetmacro{\blueComponent}{2*(50-#1)}
        \edef\clrmacro{\noexpand\cellcolor{lightblue!\blueComponent!white}}\clrmacro{\colcellbuffer#1\colcellbuffer}
    \fi
}

\newcommand{\colcellbold}[1]{%
    \ifnum #1>50
        \pgfmathsetmacro{\redComponent}{2*(#1-50)}
        \edef\clrmacro{\noexpand\cellcolor{lightred!\redComponent!white}}\clrmacro{\colcellbuffer\textbf{#1}\colcellbuffer}
    \else
        \pgfmathsetmacro{\blueComponent}{2*(50-#1)}
        \edef\clrmacro{\noexpand\cellcolor{lightblue!\blueComponent!white}}\clrmacro{\colcellbuffer\textbf{#1}\colcellbuffer}
    \fi
}

\usepackage{cog2024_anonymization}
\def\BibTeX{{\rm B\kern-.05em{\sc i\kern-.025em b}\kern-.08em
    T\kern-.1667em\lower.7ex\hbox{E}\kern-.125emX}}

\newif\ifsupplemental
\supplementalfalse 

\newcommand{\supptableref}[1]{%
  \ifsupplemental
    \ (Supplemental Table~\ref{#1})%
  \fi
}

\newcommand{\suppfigref}[1]{%
  \ifsupplemental
    \ (Supplemental Figure~\ref{#1})%
  \fi
}

\newcommand{\mapname}[1]{#1} 

\newcommand{\tablemapname}[1]{#1} 

\newcounter{suppfigure}

\newenvironment{suppfigure}
  {\renewcommand{\figurename}{Supplemental Fig.}\setcounter{figure}{\value{suppfigure}}\addtocounter{suppfigure}{1}\begin{figure}}
  {\end{figure}\setcounter{suppfigure}{\value{figure}}}

\newcounter{supptable}

\newenvironment{supptable}
  {\renewcommand{\tablename}{Supplemental Table}\setcounter{table}{\value{supptable}}\addtocounter{supptable}{1}\begin{table}}
  {\end{table}\setcounter{supptable}{\value{table}}}

\begin{document}

\title{A Competition Winning Deep Reinforcement Learning Agent in microRTS}

\author{
    \IEEEauthorblockN{\authorName}
    \IEEEauthorblockA{\authorSecondary}
}

\maketitle
\IEEEoverridecommandlockouts
\IEEEpubid{\makebox[\columnwidth]{ 979-8-3503-5067-8/24/\$31.00~\copyright2024 IEEE \hfill} 
\hspace{\columnsep}\makebox[\columnwidth]{ }}
\IEEEpubidadjcol

\begin{abstract}
Scripted agents have predominantly won the five
previous iterations of the IEEE microRTS ($\mu$RTS) competitions hosted at CIG and
CoG. Despite Deep Reinforcement Learning (DRL) algorithms making significant strides
in real-time strategy (RTS) games, their adoption in this primarily academic
competition has been limited  due to the considerable training resources required and the complexity
inherent in creating and debugging such agents. \agentName\ is the first DRL agent
to win the IEEE microRTS competition. In a benchmark without performance
constraints, \agentName\ regularly defeated the two
prior competition winners. This first competition-winning DRL submission can be
a benchmark for future microRTS competitions and a starting point for future DRL
research. Iteratively fine-tuning the base policy and transfer learning to specific maps were 
critical to \agentName's winning performance. These strategies can be used to
economically train future DRL agents. Further work in Imitation Learning using Behavior Cloning and
fine-tuning these models with DRL has proven promising as an efficient way
to bootstrap models with demonstrated, competitive behaviors.
\end{abstract}


\section{Introduction}
Deep reinforcement learning (DRL) has proven to be powerful at solving complex
problems requiring several steps to achieve a goal, such as Atari games \cite{DBLP:journals/corr/MnihKSGAWR13}, continuous
control tasks \cite{DBLP:journals/corr/LillicrapHPHETS15}, and even real-time strategy
(RTS) games like StarCraft II \cite{Vinyals2019GrandmasterLI}. The StarCraft II
grandmaster agent AlphaStar was trained with thousands of
CPUs and GPUs/TPUs for several weeks. RTS games are particularly challenging for DRL for
several reasons:
\begin{inparaenum}[(1)]
    \item the observation and action spaces are large and varied with different terrain and
        unit types;
    \item each unit type can have different actions and abilities;
    \item each action can control several units at once;
    \item rewards are sparse (win, loss, or tie) and delayed by possibly several
    thousand timesteps;
    \item winning requires combining tactical (micro) and strategic (macro) decisions;
    \item actions must be computed within a reasonable time window;
    \item the agent might not have full visibility of the game state (i.e., fog of war); and
    \item events in the game might be non-deterministic.
\end{inparaenum}

microRTS (stylized as $\mu$RTS) is a minimalist, open-source, two-player, zero-sum RTS game testbed designed for research
purposes \cite{DBLP:conf/aiide/Ontanon13}. It includes many aspects of RTS games, simplified: different unit types, unit-specific
actions, terrain, resource collection and utilization to build units, and unit-to-unit combat
where units have different strengths and weaknesses. microRTS also supports fog of war
and non-determinism; however, these were disabled for the IEEE-CoG 2023 microRTS
competition.

\begin{figure}[t]
    \centering
    \begin{tabular}{cc}
        \includegraphics[width=0.45\linewidth]{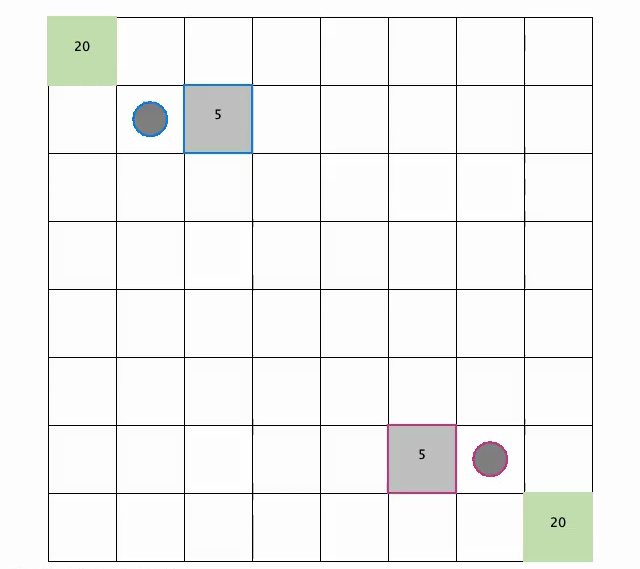} & \includegraphics[width=0.45\linewidth]{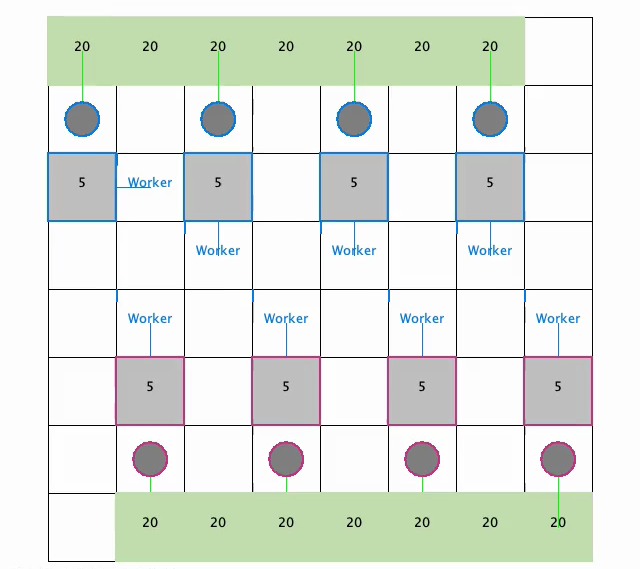} \\
        basesWorkers8x8A & FourBasesWorkers8x8 \\
        \includegraphics[width=0.45\linewidth]{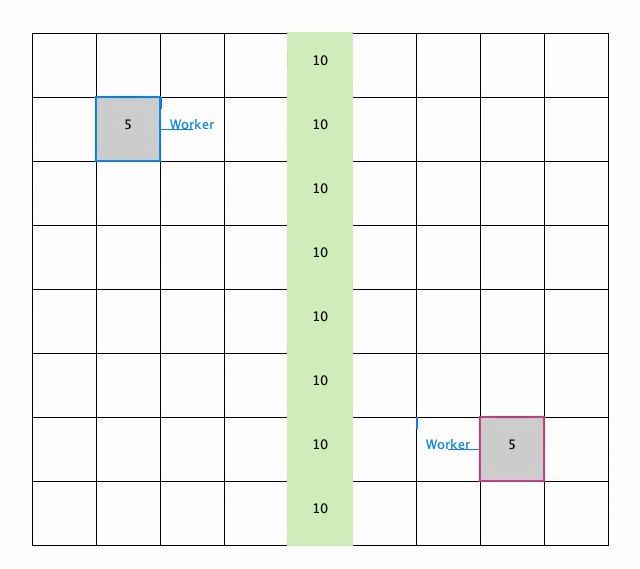} & \includegraphics[width=0.45\linewidth]{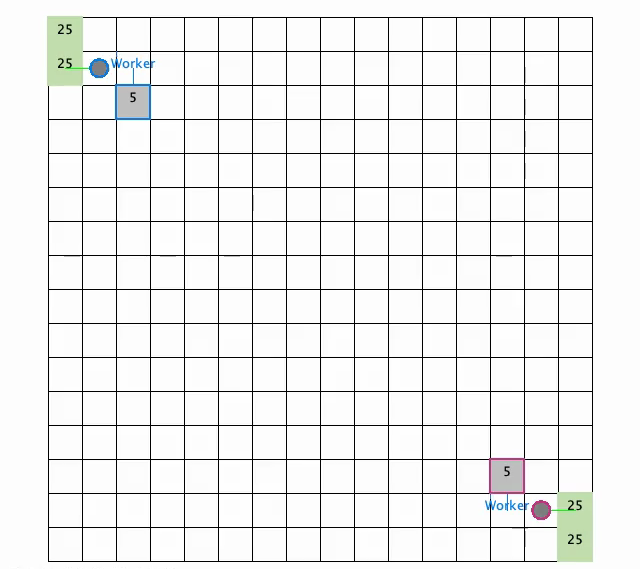} \\
        NoWhereToRun9x8 & basesWorkers16x16A \\
        \includegraphics[width=0.45\linewidth]{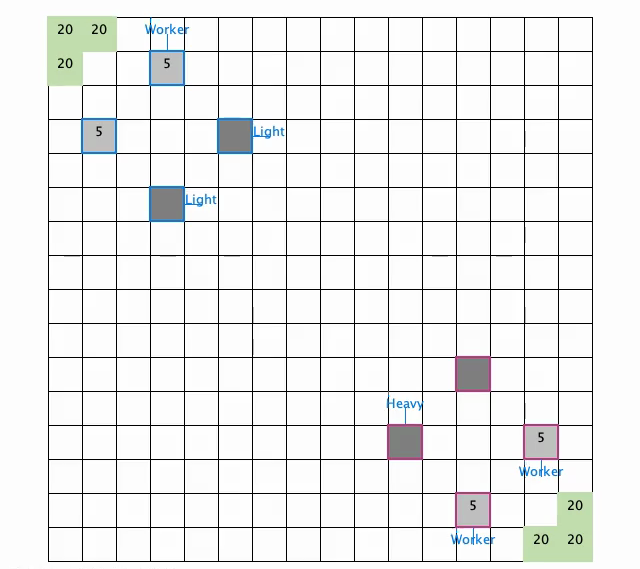} & \includegraphics[width=0.45\linewidth]{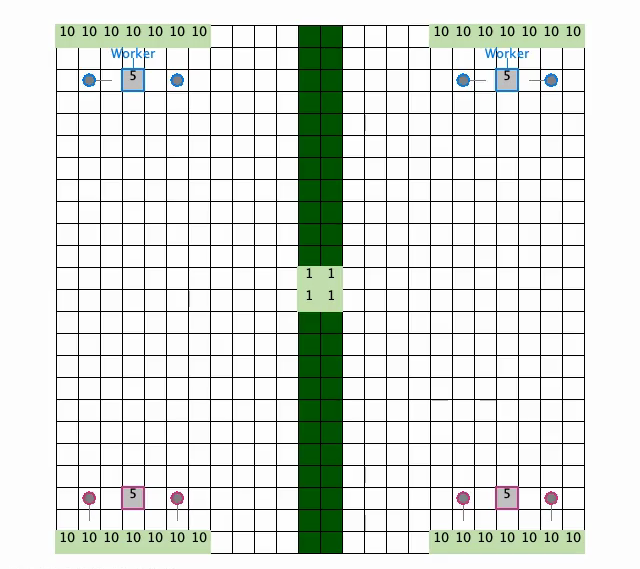} \\
        TwoBasesBarracks16x16 & DoubleGame24x24 \\
        \includegraphics[width=0.45\linewidth]{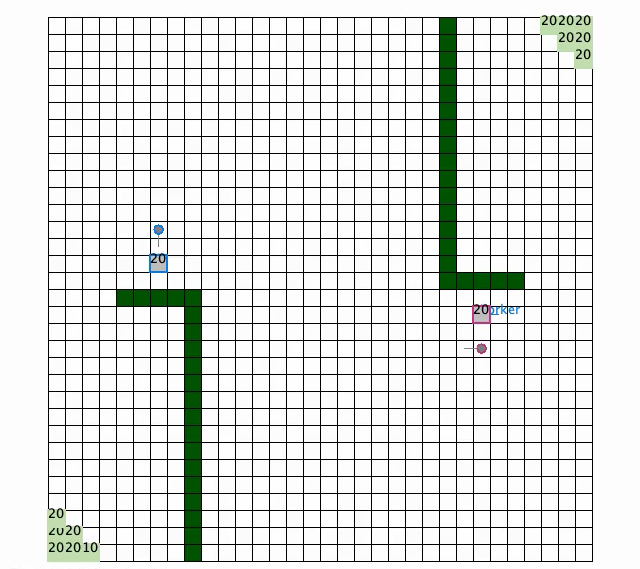} & \includegraphics[width=0.45\linewidth]{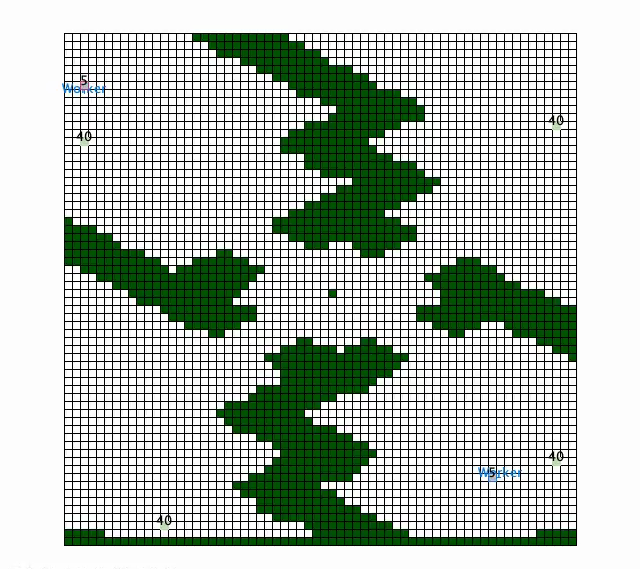} \\
        BWDistantResources32x32 & (4)BloodBath.scmB \\
    \end{tabular}
    \caption{Open competition maps.}
    \label{tab:open-maps}
\end{figure}

The IEEE microRTS competitions have been hosted at the Conference on Games (CoG) nearly
every year since 2019 and at the Conference on Computational Intelligence and Games
(CIG) before that since 2017 \cite{Ontañón_Barriga_Silva_Moraes_Lelis_2018}.
Competitors submit an agent that plays against other submissions and baselines in a round-robin tournament
on 12 different maps: 8 Open (known beforehand, Fig.~\ref{tab:open-maps}) and 4 Hidden (unknown until after the
competition results are released). Agents are supposed
to submit actions every step within 100 ms. Without GPU acceleration, this is a significant constraint for deep neural
network agents.

This paper describes how the \agentName\ agent\footnote{\raiMicroRTSGitHubUrl} was
trained and became the first DRL agent to win the microRTS competition by winning at CoG
in 2023. The agent chooses between 7 policy networks based on the map and compute
capabilities. The significant training time (70 GPU-days) combined with the general
difficulty in debugging and fine-tuning a DRL implementation could explain why DRL
hasn't been competitive so far. We demonstrate that transfer learning to specific maps
was critical to winning the competition. This strategy and our training framework can be
a starting point for future research and competition agents.

While microRTS doesn't support human players, several competition submission agents are
available to use for imitation learning. Work following the
competition shows that behavior cloning and fine-tuning with DRL can be used to train a
competitive agent more economically. Using the same playthroughs to train the critic
heads on win-loss rewards means that DRL can be trained with just sparse win-loss
rewards, eliminating the need for a handcrafted reward function.

\section{Related Work}
\subsection{MicroRTS-Py}
Reference~\cite{DBLP:journals/corr/abs-2105-13807} released
MicroRTS-Py\footnote{\url{https://github.com/Farama-Foundation/MicroRTS-Py}}, an OpenAI
Gym wrapper for microRTS that includes a Proximal Policy Optimization (PPO)
\cite{DBLP:journals/corr/SchulmanWDRK17} implementation trained on 1 of the Open maps
(they used \mapname{16x16basesWorkers}, which is the same as
\mapname{basesWorkers16x16A}). They added action composition, a shaped reward function,
invalid action masking, IMPALA-CNN \cite{DBLP:conf/icml/EspeholtSMSMWDF18} (a
convolutional neural network with residual blocks), and trained against a diverse set of
scripted agents. The agent achieved a 91\% win rate on a single map against a diverse
set of competition bots.

In their ablation studies, they found invalid action masking was essential to training a
competitive agent (82\% win rate with invalid action masking, 0\% without). Using the
residual block network IMPALA-CNN architecture instead of the Atari Nature CNN by
\cite{DBLP:journals/corr/MnihKSGAWR13} got the win rate up the rest of the way.

They experimented with two different ways to issue player actions: Unit Action
Simulation (UAS) and GridNet \cite{DBLP:conf/icml/HanSDXWSLZ19}. UAS calls the policy
iteratively on each unit, simulating the game state after each unit action before
combining all actions to submit to the game engine. GridNet computes the actions for all
units in a single policy call by computing unit action logits for all grid positions and
using a player action mask to ignore cells that don't have any units owned by the
player. UAS performed better than GridNet (91\% vs 89\%). Despite UAS's better
performance, the MicroRTS-Py library is deprecating UAS in favor of GridNet because of
UAS's more complex implementation and difficulty to incorporate self-play and imitation
learning, both features important in \agentName\ and our further work.

They tried training with self-play where the policy plays against itself, but found it
didn't improve win rate. However, we found a bug where resources
(which should be unowned) were being counted as owned by the opponent if the agent was
the second player, which likely contributed to their finding no improvement. We reimplement much of MicroRTS-Py and extend its capabilities to support training on
more maps, extend training capabilities, fix
self-play\footnote{\unownedFixGitHubCommit}, and add imitation learning.

\subsection{DeepMind's AlphaStar}
Reference~\cite{Vinyals2019GrandmasterLI}'s AlphaStar is a grandmaster-level AI trained
with DRL to play the RTS game StarCraft II. They created an initial set of agents
through imitation learning: supervised learning using a dataset of observations and
actions to train a policy to mimic the actions from the dataset. The dataset was created
by sampling replays of top-quartile human players. The supervised agents were rated in
the top 16\% of human players and used as starting points for DRL. They created a
league-based framework to train multiple agents in parallel, each with different
opponents to beat, thus creating a diverse set of training agents. AlphaStar was trained
on 3072 TPU cores and 50,400 preemptible CPU cores for a duration of 44 days.

microRTS is a much simpler game than StarCraft II, both in game mechanics and simulation
cost. Reference~\cite{DBLP:journals/corr/abs-2105-13807} trained on a single map for 300 million
steps in less than 3 GPU-days. \agentName\ trained on 10 maps for 1.5 billion steps in
70 GPU-days. While \agentName\ didn't use supervised learning to bootstrap the agent,
our following work uses imitation learning to train a competitive agent.

AlphaStar's observation and action space is significantly different from microRTS. The
StarCraft II Learning Environment (PySC2) is made to be similar to a human player's
observations and controls. An AlphaStar action is 
\begin{inparaenum}[(1)]
    \item selecting an action type,
    \item selecting a subset of units to perform the action on, and
    \item selecting a target for the action (either a map location or visible unit).
\end{inparaenum}
Once supplied an action and a target, units will perform the action until the action is complete or the unit is interrupted.
microRTS requires the agent to give single step actions for each unit at each timestep.

\subsection{Lux AI Kaggle Competitions}
The competition platform Kaggle hosts Simulation competitions where competitors submit
agents that play against other submitted agents in a turn-based game environment. Since
2020, Kaggle has featured 4 RTS-like Simulation competitions: Halite, Lux AI, Kore, and
Lux AI Season 2. Rules-based agents won Halite, Kore, and Lux AI Season 2. The first
season Lux AI winning DRL agent by \cite{lux-ai-2021-winner} has many similarities to
MicroRTS-Py: GridNet action space, reward shaping, and an actor-critic training
algorithm (IMPALA with additional UPGO and TD($\lambda$) loss terms, instead of PPO).
Instead of training with a shaped reward function throughout training,
\cite{lux-ai-2021-winner} used shaped rewards on a smaller map before transitioning to
sparse win-loss rewards on larger and competition-size maps. The top DRL agent by
\cite{Ferdinand2021doublecone} in the Lux AI Season 2 competition used a "DoubleCone"
neural network backbone with critic and actor heads. DoubleCone is similar to ResNet's
backbone but the middle residual blocks are downscaled 4x to reduce inference time.
\agentName\ transitions from shaped to sparse rewards during training and uses the
DoubleCone architecture.

\section{Methods}
\label{sec:methods}

Compared to \cite{DBLP:journals/corr/abs-2105-13807}, \agentName's biggest differences
are 
\begin{inparaenum}[(1)]
    \item the DoubleCone neural network architecture,
    \item adding self-play to the training regime,
    \item using a training schedule that transitions from shaped to sparse rewards,
    \item training on multiple maps, and
    \item using transfer learning to train specialized models for specific maps.
\end{inparaenum}
\agentName\ loads 7 policy networks, but only uses one network at a time chosen based
on map and compute capabilities (Table~\ref{tab:policy-networks}).
Networks are selected by
\begin{inparaenum}[(1)]
    \item gathering all networks that are compatible with the map and its size,
    \item prioritizing map-specific networks over size-specific networks, and
    \item picking the highest priority network that can run within the allotted time on
    the current hardware.
\end{inparaenum}
microRTS supports any map size (even non-square), and observations are padded to fit the
policy network. Policy actions are clipped to fit the map size.

We reimplement much of MicroRTS-Py, including the PPO implementation, action
composition, shaped reward function, invalid action masking, GridNet, self-play, and
scripted bot training. \agentName's codebase\footnote{\rlAlgoImplsGitHubUrl} supports 
environments beyond microRTS and reimplementing allowed the environment to fit into the 
existing codebase. We extended the observation representation in two ways:
\begin{inparaenum}[(1)]
    \item walls and
    \item unit destinations as invalid move targets in the invalid action mask.
\end{inparaenum}
Only units' current locations were considered invalid move targets in MicroRTS-Py;
however, microRTS doesn't allow units to move where another unit is moving into. Padded
positions were represented as walls, which are impassable and noninteractive.

\subsection{Neural Network Architecture}
\begin{figure}[t]
    \centering
    \includegraphics[width=0.9\linewidth]{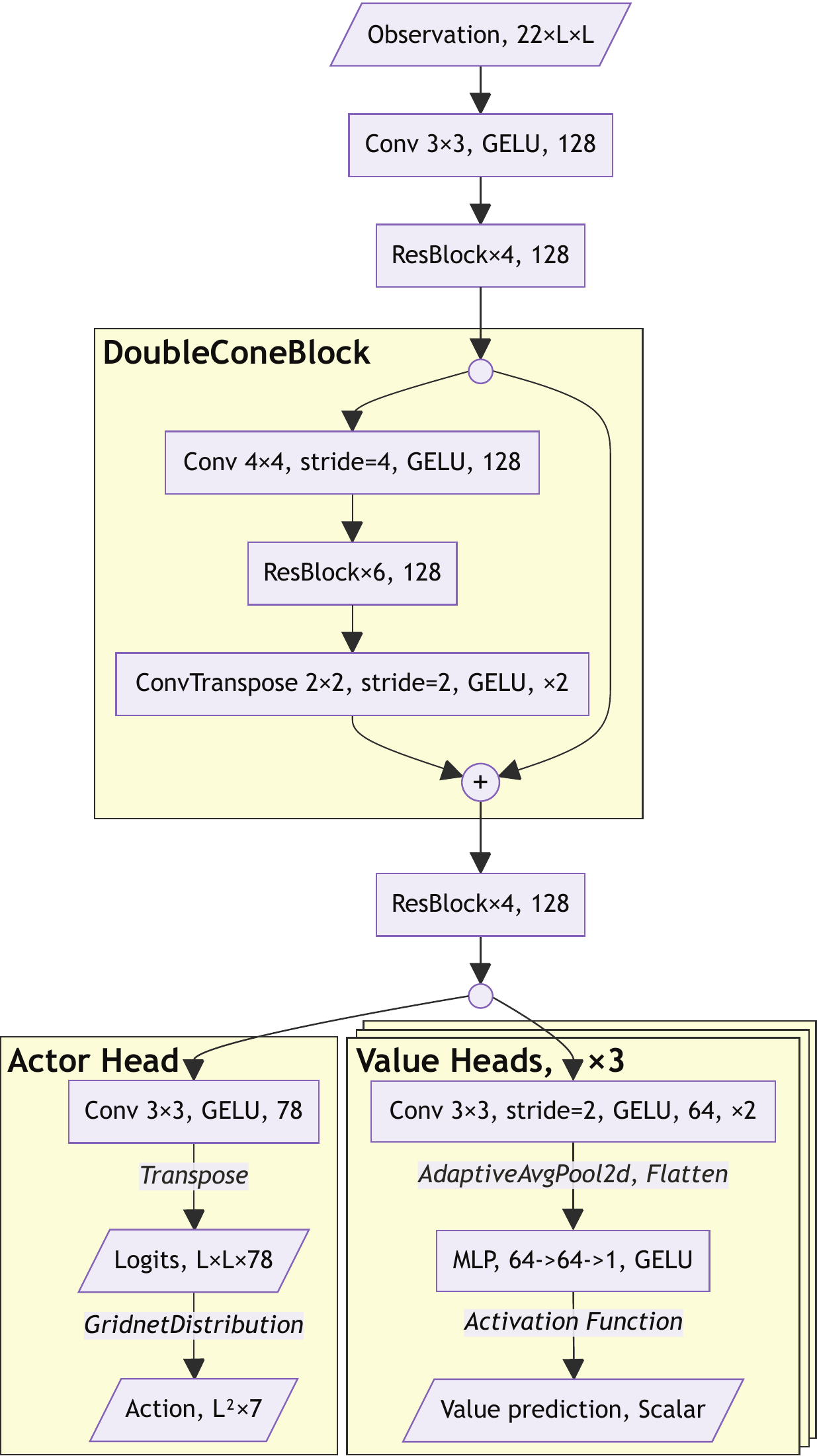}
    \caption{DoubleCone architecture.}
    \label{fig:doublecone}
\end{figure}

\agentName\ uses two different neural network backbones:
\cite{Ferdinand2021doublecone}'s DoubleCone(4, 6, 4) (Fig.~\ref{fig:doublecone}) and a
custom network (squnet). The actor head is a convolutional layer that outputs logits for
unit actions at every position. A unit action is composed of independent discrete
subactions: $D = \{a_{\text{action type}},$ $a_{\text{move direction}},$
$a_{\text{harvest direction}},$ $a_{\text{return direction}},$ $a_{\text{produce
direction}},$ $a_{\text{produce type}},$ $a_{\text{relative attack position}}\}$.
Invalid action masking sets logits to a very large negative number (thus zeroing
probabilities and gradients) for actions that are illegal or would accomplish nothing
(e.g., moving a unit to an occupied or reserved position). This masking significantly
reduces the action space per turn and makes training more efficient
\cite{DBLP:journals/corr/abs-2006-14171}.

While DoubleCone could support any map size, inference would likely exceed 100
milliseconds for larger maps. Therefore, for larger maps, we used a different
architecture (squnet) which nests 3 downscaling blocks\suppfigref{fig:squnet}. This 
creates a network shaped like U-Net, but functionally similar to DoubleCone.
This aggressive downscaling reduces the number of operations necessary during inference,
especially for larger maps where squnet does 8-fold fewer operations compared to
DoubleCone on the largest 64x64 map\supptableref{table:architectureBreakdown}.

Instead of 1 value head, \agentName\ uses 3 values heads for 3 different value
functions:
\begin{inparaenum}[(1)]
    \item shaped reward similar to MicroRTS-Py except each combat unit type is
    scaled by build-time (rewarding expensive units more),
    \item win-loss sparse reward at game end (Tanh activation), and
    \item in-game difference in units based on cost (similar to the reward function used
    by \cite{Winter2021}).
\end{inparaenum}
These 3 value heads are used to mix-and-match rewards over the course of training,
generally weighing the dense reward heads (1 and 3) more heavily at the start of
training and finishing with only the win-loss reward contributing at the end.

\subsection{Base Model Training}
We train using the PPO loss function from \cite{DBLP:journals/corr/SchulmanWDRK17}:
\begin{align}
    L(\theta) &= \hat{\mathbb{E}}_t \left[ L^{\text{CLIP}}(\theta) - c_1 L^{\text{VF}}(\theta) + c_2 S\left[\pi_\theta\right](s_t) \right], \\
    L^{\text{CLIP}}(\theta) &= \min \bigg( \frac{\pi_\theta(a_t|s_t)}{\pi_{\theta_{old}}(a_t|s_t)} \hat{A}_t, \notag \\
    &\qquad \text{clip}\left( \frac{\pi_\theta(a_t|s_t)}{\pi_{\theta_{old}}(a_t|s_t)}, 1-\epsilon, 1+\epsilon \right) \hat{A}_t \bigg), \\
    L^{\text{VF}}(\theta) &= \frac{1}{2} \left( V_\theta(s_t) - \hat{V}_t \right)^2,
\end{align}
where $c_1$ is the value loss coefficient, $c_2$ is the entropy coefficient, $S$ is an
entropy bonus function, $\pi_\theta$ is the stochastic policy, $\pi_{\theta_{old}}$ is the rollout policy, $a_t$
is the action taken at time $t$, $s_t$ is the observation at time $t$, $\hat{A}_t$
is the advantage estimate,  $\epsilon$ is the clipping coefficient, $V_\theta$ is the
value function, $\hat{V}_t$ is the return estimate. We estimate the advantage
using Generalized Advantage Estimation (GAE) \cite{DBLP:journals/corr/SchulmanMLJA15}.
We compute a separate advantage for each of the 3 rewards with independent $\gamma$
(discount factor) and $\lambda$ (exponential weight discount) for each reward. The 3 
advantages are weighted summed together to get the final advantage estimate, which is
used in computing the policy loss ($L^{\text{CLIP}}$).

The reward weights, value loss coefficients (each value
head has its own loss coefficient), entropy coefficient, and learning rate are varied on
a schedule\supptableref{tab:initial-training-schedule}. At the start of training, the policy loss heavily weighs the shaped 
reward advantage, and the value loss similarly weighs towards the shaped value head. By the end
of training, both losses are weighted more towards the win-loss sparse reward. The entropy coefficient is also
lowered at the end of training to discourage the agent from making as many random moves as the
learning rate is lowered. The schedule specifies phases when these values are set and
transitions between phases by changing values linearly based on timesteps.

We first trained \agentName\ with self-play both against itself and against prior
versions of itself\supptableref{tab:training-parameters}. We reached a 91\% win rate against the same bots MicroRTS-Py was
benchmarked against. However, it only beat CoacAI (the 2020 competition winner) in 20\%
of games. The best performing agent of MicroRTS-Py nearly always beat CoacAI; however,
the best versions using GridNet also usually lost against CoacAI. We fine-tuned
the model through 3 iterations:
\begin{inparaenum}[(1)]
    \item one-half of environments trained against CoacAI\supptableref{tab:shaped-finetuning-schedule};
    \item one-half of environments trained against CoacAI or Mayari (2021 competition
    winner) split evenly and primarily trained on win-loss rewards\supptableref{tab:sparse-finetuning-schedule}; and
    \item same as before with action mask improvements and a GELU activation
    after the stride-4 convolution to match \cite{Ferdinand2021doublecone}'s DoubleCone.
\end{inparaenum}
By the end of fine-tuning, the model was winning 98\% of games, including about 90\%
against each of CoacAI and Mayari.

\subsection{Transfer Learning}
Up to this point, the model had only been trained on the 5 smaller Open maps. Using the
fine-tuned model parameters to jumpstart training \cite{DBLP:books/sp/12/Lazaric12}, we
trained additional models, each trained exclusively on 1 of 3 Open maps.
\mapname{NoWhereToRun9x8} is very different from the other smaller maps with a wall of
resources separating opponents. \mapname{DoubleGame24x24} and
\mapname{BWDistantResources32x32} are larger than the maps the base model trained on.
All 3 transfer learning runs used the same schedule:
\begin{inparaenum}[(1)]
    \item warmup of sparse, win-loss reward weights linearly transitioned to a mixture
    of both shaped and sparse rewards,
    \item middle phase of mixed rewards,
    \item end phase of sparse rewards at a lower learning rate\supptableref{tab:transfer-learning-schedule}.
\end{inparaenum}
We did additional fine-tuning on \mapname{NoWhereToRun9x8} using only the win-loss
reward. These transfer learned agents exceeded 90\% win rate on their respective maps,
significant improvements over the base model (especially on,
\mapname{BWDistantResources32x32} which started below 10\%).

\subsection{Squnet Training}
We trained the squnet models with fewer steps because of time constraints and mostly
didn't use the cost-based reward because it didn't help train the base
model\supptableref{tab:squnet-training-parameters}. The 2 models (trained on maps of up
to size 32x32 and 64x64, respectively) managed only a 40\% win rate, never beating
CoacAI or Mayari. These models were policies of last resort.

We also fine-tuned the squnet-map32 model on only \mapname{BWDistantResources32x32}
using the sparse reward fine-tuning schedule. This fine-tuned 
model achieved 85\% win rate, beating Mayari half the time, but never beating CoacAI.

\subsection{Behavior Cloning Bootstrapped Training}
In follow-up work after the \agentName\ submission, we wanted to train a model that
\begin{inparaenum}[(1)]
    \item didn't require the shaped rewards and reward scheduling,
    \item could be trained in fewer steps and less time, and
    \item could defeat prior competition winners on the largest maps.
\end{inparaenum}
We opted for a neural architecture between DoubleCone and squnet: a nested downscaling residual
block each of stride 4, so the bottom block scales the input down to 1/16th the original
size. At each downscaling level, there were multiple residual blocks (6 at full
resolution, split evenly by the downscaling block; 4 at 1/4 resolution, split evenly by
the 1/16 downscaling block; and 4 at 1/16 resolution)\supptableref{tab:bc-architecture}. This architecture theoretically
has a 128x128 receptive field while using 25\% fewer operations than DoubleCone at inference time. On
the largest Open map (\mapname{(4)BloodBath.scmB}), this is 6 times more
computations than squnet-64. Therefore, this neural architecture wouldn't be usable in a
competition given the same hardware constraints as the 2023 competition.

Initially, we tried a similar training strategy to \agentName\ where the model is
trained on 16x16 maps and that model is used for transfer learning to
larger maps. However, we only managed a 60\% win rate on
\mapname{BWDistantResources32x32} and less than a 20\% win rate on
\mapname{(4)BloodBath.scmB} after over 100 million
steps before terminating training.

Next, we tried imitation learning to bootstrap the model, similar to
\cite{Vinyals2019GrandmasterLI}. We got rid of the three rewards, opting for only the
win-loss reward. microRTS doesn't have human replays, so we used playthroughs of the
2021 competition winner Mayari playing against itself, 2020 competition winner CoacAI, 
and POLightRush (baseline scripted bot and 2017 competition
winner)\supptableref{tab:bc-training-parameters}.
Instead of generating an offline replay dataset, we set the microRTS environment to play bots
against each other and these observations and actions were fed into rollouts used for
behavior cloning the policy and fitting the value heads:
\begin{align}
    L^{\text{BC+VF}}(\theta) &= \hat{\mathbb{E}}_t \left[ L^{\text{BC}}(\theta) + c_1 L^{\text{VF}}(\theta) \right], \\
    L^{\text{BC}}(\theta) &= -\frac{1}{\|a_t\|} \log \pi_{\theta}(a_t|s_t),
\end{align}
where $c_1$ and $L^{\text{VF}}$ are the same as in the PPO loss function. The behavior cloning
policy loss is the cross-entropy loss between the policy logits and the actions taken by
the Mayari bot. We found scaling the loss by the number of units accepting actions
allowed the learning rate to be significantly increased. Scaling down the loss by the
number of units keeps the losses for all turns roughly similar in scale as otherwise
large unit count turns would have much larger losses as each unit's loss contribution is
summed together (each unit's actions are assumed to be independent).

We trained 3 behavior cloned models (16x16, 32x32, and 64x64) on the same maps for each
map size as \agentName\ training. The 64x64 model used the weights of the 32x32 model 
as a starting point, while the other two models were randomly initialized. We then used 
PPO to fine-tune the behavior cloned models on the same maps\supptableref{tab:bc-ppo-training-parameters}.

\section{Results}
\subsection{Single Player Round-robin Benchmark}
\label{sec:single-player-benchmark}
In a single player round-robin benchmark on the Open maps (Table~\ref{tab:single-player-winrate}), \agentName\
beat the competition winners of 2021 (Mayari), 2020 (CoacAI), and 2017 (POLightRush,
baseline) on 7 of the 8 maps (winning over 96\% of games on these maps). \agentName\
could only beat the POWorkerRush baseline bot on the largest map,
\mapname{(4)BloodBath.scmB}. The DistantResources fine-tuned squnet model performed worse than the DoubleCone model
across all opponents, but maintained an over 50\% win rate against all but CoacAI. Timeouts didn't affect results significantly.

\begin{table}[t]
    \centering
    \caption{Single player round-robin benchmark win rates. Win rates over 50\% are bolded. Higher win rates are redder. Lower win rates are bluer.}
    \label{tab:single-player-winrate}
    \begin{threeparttable}
    \arrayrulecolor{black}
    \begin{tabular}{lcccc|c}
     & \rotatebox{90}{\textbf{WorkerRush}} & \rotatebox{90}{\textbf{LightRush}} & \rotatebox{90}{\textbf{CoacAI}} & \rotatebox{90}{\textbf{Mayari}} & \rotatebox{90}{\textbf{Overall}} \\
    \arrayrulecolor{black}\specialrule{.5pt}{0pt}{0pt}
    basesWorkers8x8A & \colcell{95} & \colcell{100} & \colcell{99} & \colcell{100} & \colcell{99} \\
    FourBasesWorkers8x8 & \colcell{100} & \colcell{100} & \colcell{100} & \colcell{98} & \colcell{100} \\
    NoWhereToRun9x8 & \colcell{100} & \colcell{100} & \colcell{93} & \colcell{99} & \colcell{98} \\
    basesWorkers16x16A & \colcell{100} & \colcell{100} & \colcell{90} & \colcell{98} & \colcell{97} \\
    TwoBasesBarracks16x16 & \colcell{100} & \colcell{89} & \colcell{99} & \colcell{100} & \colcell{97} \\
    DoubleGame24x24 & \colcell{100} & \colcell{98} & \colcell{94} & \colcell{100} & \colcell{98} \\
    BWDistantResources32x32\tnote{a} & \colcell{99}/\textbf{93} & \colcell{90}/\textbf{73} & \colcell{88}/23 & \colcell{99}/\textbf{58} & \colcell{94}\tnote{b}{ }/\textbf{61} \\
    (4)BloodBath.scmB & \colcell{98} & \colcell{0} & \colcell{0} & \colcell{0} & \colcell{25}\tnote{c} \\
    \arrayrulecolor{black}\specialrule{.5pt}{0pt}{0pt}
    AI Average\tnote{d} & \colcell{99} & \colcell{85} & \colcell{83} & \colcell{87} & \colcell{88} \\
    \end{tabular}
    \begin{tablenotes}
    \item[a] Two models were trained for BWDistantResources32x32: DoubleCone (1\textsuperscript{st} number)
    and squnet (2\textsuperscript{nd} number). Cell color based on DoubleCone results.
    \item[b] \agentName\ lost 0.25\% of matches (1 match) by timeout.
    \item[c] \agentName\ lost  1\% of matches (4 matches) by timeout.
    \item[d] AI Average uses the DoubleCone (1\textsuperscript{st} number) results from BWDistantResources32x32.
    \end{tablenotes}
    \end{threeparttable}
\end{table}

\subsection{IEEE-CoG 2023 microRTS Competition Results}
The IEEE-CoG 2023 microRTS competition is a round-robin tournament on 12 maps of
different sizes and distributions of terrain, resources, and starting units and
buildings. 8 Open maps are known beforehand, 4 Hidden maps are only revealed after the
competition. The winner is the agent with the highest win rate on the 8 Open maps.
Hidden map results are publicly available, but this paper will only discuss
the Open maps. For
this competition, a total of 11 agents were submitted: 9 programmatic policies, 1
synthesized programmatic policy, and \agentName. The competition also had 6
baselines:
\begin{inparaenum}[(1)]
    \item RandomBiasedAI (performs actions randomly, biased towards attacking if able),
    \item NaiveMCTS (a simple Monte Carlo tree search agent that searches until reaching
    the time limit)
    \item POWorkerRush,
    \item POLightRush,
    \item 2L (programmatic strategies generated by \cite{DBLP:conf/ijcai/MoraesAFL23}, the
    competition organizers), and
    \item the prior competition winner Mayari.
\end{inparaenum}
The baselines cannot win the competition.

\agentName\ was declared the winner with the highest win rate (72\%) across all
submissions (Table~\ref{tab:competition-winrate}). \agentName\ had a higher win rate 
than all but two baselines: \texttt{2L} (76\%) and Mayari (82\%). \agentName\ had an 
over 50\% win rate versus every opponent including \texttt{2L} (60\%) and Mayari (65\%).

\begin{table}[t]
    \centering
    \caption{Win rates of selection of agents in the IEEE-CoG 2023 microRTS competition.
    The row agent is player 1, while the column agent is player 2. The win rate value is
    the percentage of games won by player 1. Cells are bolded if the win
    rate is higher than the opponent's row win rate.}
    \label{tab:competition-winrate}
    \begin{threeparttable}
    \arrayrulecolor{black}
    \begin{tabular}{lcccccc|c}
    & \begin{sideways} Mayari \end{sideways} 
    & \begin{sideways} 2L \end{sideways} 
    & \begin{sideways} \textbf{\agentName} \end{sideways} 
    & \begin{sideways} ObiBotKenobi \end{sideways} 
    & \begin{sideways} POLightRush \end{sideways} 
    & \begin{sideways} POWorkerRush \end{sideways} 
    & \begin{sideways} Overall\tnote{a} \end{sideways} \\
    \arrayrulecolor{black}\specialrule{.5pt}{0pt}{0pt}
    Mayari (2021 winner) & - & \colcellbold{53} & \colcellnobold{32} & \colcellbold{73} & \colcellbold{88} & \colcellbold{75} & \colcellnobold{82} \\
    2L (baseline) & \colcellnobold{51} & - & \colcellnobold{39} & \colcellbold{50} & \colcellbold{75} & \colcellbold{88} & \colcellnobold{76} \\
    \textbf{\agentName} (2023 winner) & \colcellbold{62} & \colcellbold{59} & - & \colcellbold{49}\tnote{b} & \colcellbold{64} & \colcellbold{78} & \colcellnobold{72} \\
    ObiBotKenobi (2023 2\textsuperscript{nd} place) & \colcellnobold{39} & \colcellnobold{29} & \colcellnobold{47} & - & \colcellbold{58} & \colcellbold{65} & \colcellnobold{66} \\
    POLightRush (baseline) & \colcellnobold{0} & \colcellnobold{25} & \colcellnobold{29} & \colcellnobold{38} & - & \colcellbold{69} & \colcellnobold{55} \\
    POWorkerRush (baseline) & \colcellnobold{13} & \colcellnobold{13} & \colcellnobold{21} & \colcellnobold{29} & \colcellnobold{38} & - & \colcellnobold{53} \\
    \end{tabular}
    \begin{tablenotes}
    \item[a] Overall includes all agents, including those not shown.
    \item[b] \agentName\ vs ObiBotKenobi is bolded because 49\% is higher than 47\%, thus meaning the combined player 1
    and 2 win rate is 51\% for \agentName\ vs ObiBotKenobi.
    \end{tablenotes}
    \end{threeparttable}
\end{table}

As expected from the single player round-robin benchmark, \agentName\ does better on
smaller maps and dismally on the largest maps
(Table~\ref{tab:competition-winrate-by-map}). However, in the competition, \agentName\
underperformed against agents already benchmarked in the single player round-robin
(14-19\% lower win rate against each agent), even accounting for the likely use of
the weaker squnet model on \mapname{BWDistantResources32x32}.  Breaking down by map, 
\agentName\ underperformed against benchmarked agents by 20-40\% on 5 maps.

\begin{table}[t]
    \centering
    \caption{\agentName\ win rates in 2023 competition by opponent and map. Win rates over 50\% are bolded. Higher win rates are redder. Lower win rates are bluer.}
    \label{tab:competition-winrate-by-map}
    \begin{threeparttable}
    \arrayrulecolor{black}
    \begin{tabular}{lccccc|c}
    & \rotatebox{90}{POWorkerRush} & \rotatebox{90}{POLightRush} & \rotatebox{90}{ObiBotKenobi} & \rotatebox{90}{2L} & \rotatebox{90}{Mayari} & \rotatebox{90}{Overall\tnote{a}}\\ 
    \arrayrulecolor{black}\specialrule{.5pt}{0pt}{0pt}
    basesWorkers8x8A & \colcell{60} & \colcell{70} & \colcell{60} & \colcell{60} & \colcell{60} & \colcell{66}\\ 
    FourBasesWorkers8x8 & \colcell{100} & \colcell{100} & \colcell{20} & \colcell{95} & \colcell{100} & \colcell{95}\\ 
    NoWhereToRun9x8 & \colcell{90} & \colcell{85} & \colcell{83} & \colcell{70} & \colcell{70} & \colcell{84}\\
    basesWorkers16x16A & \colcell{100} & \colcell{100} & \colcell{95} & \colcell{100} & \colcell{100} & \colcell{100}\\
    TwoBasesBarracks16x16 & \colcell{80} & \colcell{80} & \colcell{10} & \colcell{70} & \colcell{80} & \colcell{75}\\
    DoubleGame24x24 & \colcell{80} & \colcell{75} & \colcell{78} & \colcell{80} & \colcell{75} & \colcell{80}\\ 
    BWDistantResources32x32 & \colcell{50} & \colcell{30} & \colcell{35} & \colcell{3} & \colcell{35} & \colcell{54}\\
    (4)BloodBath.scmB & \colcell{70} & \colcell{0} & \colcell{28} & \colcell{0} & \colcell{0} & \colcell{34}\\ 
    \arrayrulecolor{black}\specialrule{.5pt}{0pt}{0pt}
    AI Average & \colcell{79} & \colcell{68} & \colcell{51} & \colcell{60} & \colcell{65} & \colcell{74}\\ 
    \end{tabular}
    \begin{tablenotes}
    \item[a] Overall includes all agents, including those not shown.
    \end{tablenotes}
    \end{threeparttable}
\end{table}

The competition ran jobs splitting each map into 5 or 10 jobs where
each job would run a complete round-robin with all agents on that map playing 2 or 1
games, respectively, as player 1 and 2 each. For \mapname{basesWorkers8x8A}, on which \agentName\
underperformed by almost 40\%, the competition had 5 jobs. On the first 3 jobs,
\agentName\ won nearly every game. On the last 2 jobs, \agentName\ lost nearly every
game. 1 or 2 jobs per underperforming map appear to have outlier low win rates
for \agentName\ (Table~\ref{tab:outlier-winrate}).

\begin{table}[t]
    \centering
    \caption{\agentName\ win rates split by competition job. Outlier jobs are bolded.}
    \label{tab:outlier-winrate}
    \begin{threeparttable}
    \begin{tabular}{r|rrrrrrrr|l}
    & \begin{sideways}\tablemapname{basesWorkers8x8A}\end{sideways} &
    \begin{sideways}\tablemapname{FourBasesWorkers8x8}\end{sideways} &
    \begin{sideways}\tablemapname{NoWhereToRun9x8}\end{sideways} &
    \begin{sideways}\tablemapname{basesWorkers16x16A}\end{sideways} &
    \begin{sideways}\tablemapname{TwoBasesBarracks16x16}\end{sideways} &
    \begin{sideways}\tablemapname{DoubleGame24x24}\end{sideways} &
    \begin{sideways}\tablemapname{BWDistantResources32x32}\end{sideways} &
    \begin{sideways}\tablemapname{(4)BloodBath.scmB}\end{sideways} & 
    \begin{sideways}Overall\end{sideways}\\
    \arrayrulecolor{black}\specialrule{.5pt}{0pt}{0pt}
    & \colcellnobold{98} & \colcellnobold{97} & \colcellnobold{97} & \colcellnobold{100} & \colcellnobold{94} & \colcellnobold{99} & \colcellnobold{49} & \colcellnobold{34} &  \\
    & \colcellnobold{100} & \colcellnobold{94} & \colcellnobold{97} & \colcellnobold{100} & \colcellbold{4} & \colcellnobold{96} & \colcellnobold{53} & \colcellnobold{38} &  \\
    & \colcellnobold{100} & \colcellnobold{95} & \colcellbold{33} & \colcellnobold{100} & \colcellnobold{92} & \colcellnobold{100} & \colcellnobold{54} & \colcellnobold{38} &  \\
    & \colcellbold{5} & \colcellnobold{92} & \colcellnobold{94} & \colcellnobold{98} & \colcellnobold{94} & \colcellnobold{97} & \colcellnobold{58} & \colcellnobold{39} &  \\
    & \colcellbold{28} & \colcellnobold{95} & \colcellnobold{100} & \colcellnobold{100} & \colcellnobold{94} & \colcellbold{9} & \colcellnobold{58} & \colcellbold{19} &  \\
    & \multicolumn{1}{l}{} & \multicolumn{1}{l}{} & \colcellnobold{100} & \multicolumn{1}{l}{} & \multicolumn{1}{l}{} & \multicolumn{1}{l}{} & \multicolumn{1}{l}{} & \colcellnobold{39} &  \\
    & \multicolumn{1}{l}{} & \multicolumn{1}{l}{} & \colcellnobold{97} & \multicolumn{1}{l}{} & \multicolumn{1}{l}{} & \multicolumn{1}{l}{} & \multicolumn{1}{l}{} & \colcellbold{19} &  \\
    & \multicolumn{1}{l}{} & \multicolumn{1}{l}{} & \colcellbold{30} & \multicolumn{1}{l}{} & \multicolumn{1}{l}{} & \multicolumn{1}{l}{} & \multicolumn{1}{l}{} & \colcellnobold{44} &  \\
    & \multicolumn{1}{l}{} & \multicolumn{1}{l}{} & \colcellnobold{97} & \multicolumn{1}{l}{} & \multicolumn{1}{l}{} & \multicolumn{1}{l}{} & \multicolumn{1}{l}{} & \colcellnobold{36} &  \\
    & \multicolumn{1}{l}{} & \multicolumn{1}{l}{} & \colcellnobold{95} & \multicolumn{1}{l}{} &
    \multicolumn{1}{l}{} & \multicolumn{1}{l}{} & \multicolumn{1}{l}{} & \colcellnobold{36} &  \\
    \arrayrulecolor{black}\specialrule{.5pt}{0pt}{0pt}
   Average\tnote{a} & \colcellnobold{66} & \colcellnobold{95} & \colcellnobold{84} & \colcellnobold{100} & \colcellnobold{75} & \colcellnobold{80} & \colcellnobold{54} & \colcellnobold{34} & \multicolumn{1}{r}{\colcellnobold{74}} \\
   No Outliers\tnote{b} & \colcellnobold{99} & \colcellnobold{95} & \colcellnobold{97} & \colcellnobold{100} & \colcellnobold{93} & \colcellnobold{98} & \colcellnobold{54} & \colcellnobold{38} & \multicolumn{1}{r}{\colcellnobold{84}} \\
    \end{tabular}
    \begin{tablenotes}
        \item[a] Average is the average win rate for all jobs.
        \item[b] No Outliers is the average win rate for all jobs excluding "outlier" jobs.
    \end{tablenotes}
    \end{threeparttable}
\end{table}

\subsection{Behavior Cloning Results}
\label{sec:behavior-cloning-results}
We created 2 additional agents:
\begin{inparaenum}[(1)]
    \item only behavior cloning (\bcAgent) and
    \item a PPO fine-tuned follow-up of \bcAgent\ (\bcPPOAgent).
\end{inparaenum}
Each agent consisted of the models trained on their
respective map sizes (16x16, 32x32, and 64x64). These agents do not have any
map-specific models.

\bcAgent\ had a 71\% win rate, doing well against the POLightRush baseline (96\%, better than \agentName) and respectably against
Mayari (44\%)\supptableref{tab:bc-winrate}. On the largest map, \bcAgent\ manages to occasionally beat POLightRush (63\%), CoacAI
(20\%), and Mayari (40\%) compared to \agentName's 0\% against these 3 more difficult opponents.

Once fine-tuned with PPO, \bcPPOAgent\ obtains an \agentName-comparable 88\% win rate
(Table~\ref{tab:bcppo-winrate}). \bcPPOAgent\ generally improves upon \bcAgent's win
rates on each map and against each opponent. However, the biggest exceptions are
POLightRush on \mapname{TwoBasesBarracks16x16} (from 100\% to 0\%) and the largest map
where the fine-tuned model can no longer beat CoacAI and Mayari.

\begin{table}[t]
    \centering
    \caption{\bcPPOAgent\ win rate in a single player round-robin benchmark. Win rates over 50\% are bolded. Higher win rates are redder. Lower win rates are bluer.}
    \label{tab:bcppo-winrate}
    \arrayrulecolor{black}
    \begin{tabular}{lcccc|c}
    & \rotatebox{90}{POWorkerRush} & \rotatebox{90}{POLightRush} & \rotatebox{90}{CoacAI} & \rotatebox{90}{Mayari} & \rotatebox{90}{Overall} \\
    \arrayrulecolor{black}\specialrule{.5pt}{0pt}{0pt}
    basesWorkers8x8A & \colcell{92} & \colcell{100} & \colcell{85} & \colcell{100} & \colcell{94} \\
    FourBasesWorkers8x8 & \colcell{100} & \colcell{100} & \colcell{100} & \colcell{100} & \colcell{100} \\
    NoWhereToRun9x8 & \colcell{100} & \colcell{100} & \colcell{90} & \colcell{80} & \colcell{92} \\
    basesWorkers16x16A & \colcell{100} & \colcell{100} & \colcell{95} & \colcell{95} & \colcell{98} \\
    TwoBasesBarracks16x16 & \colcell{100} & \colcell{0} & \colcell{100} & \colcell{95} & \colcell{74} \\
    DoubleGame24x24 & \colcell{98} & \colcell{85} & \colcell{100} & \colcell{100} & \colcell{96} \\
    BWDistantResources32x32 & \colcell{100} & \colcell{100} & \colcell{95} & \colcell{100} & \colcell{99} \\
    (4)BloodBath.scmB & \colcell{100} & \colcell{88} & \colcell{0} & \colcell{5} & \colcell{48} \\
    \arrayrulecolor{black}\specialrule{.5pt}{0pt}{0pt}
    AI Average & \colcell{99} & \colcell{84} & \colcell{83} & \colcell{84} & \colcell{88} \\
    \end{tabular}
\end{table}

While \agentName\ required map-specific fine-tuned models to be competitive on
\mapname{NoWhereToRun9x8}, \mapname{DoubleGame24x24}, and \mapname{BWDistantResources32x32},
\bcPPOAgent\ only has models for the different map-sizes. This demonstrates promise for
creating generalized agents that can play across a wide range of maps.

\section{Discussion}
\subsection{Improving Inference Time in microRTS Competitions}
\agentName's underperformance in the 2023 competition suggests that job
environments can run slow. We worked with the competition
organizers to reduce the chance of timeouts. However, it was difficult to reproduce the same results as
the competition servers in our development environments.

Improving inference time is critical to matching benchmark results in a competition.
We suggest 3 improvements (2 for agents, and 1 for the competition organizers):
\begin{inparaenum}[(1)]
    \item use fast inference runtime providers like OpenVINO for ONNX Runtime,
    \item train using smaller models (possibly using behavior cloning or policy
    distillation from larger trained models to bootstrap training), and
    \item replace the fixed per turn timeout tolerance in the competition with an overtime
    budget.
\end{inparaenum}
For DoubleCone, \cite{Ferdinand2021doublecone} found using OpenVINO could have made
inference 2-3 times faster in the LUX competition. This would likely make running
DoubleCone and the larger squnet models on all but the largest maps feasible for the
competition. An overtime budget for an entire match instead of the 20 ms per turn
tolerance will help agents deal with environment instabilities. For example,
\bcPPOAgent\ timed out in 11\% of games on \mapname{(4)BloodBath.scmB}, despite
averaging 55 ms/turn and going over 100 ms in only 0.016\% of turns (averaging less than
1 over 100 ms turn per game). An overage budget of even 1 second per game would likely
prevent most timeouts.

\subsection{Training on Larger Maps}
None of our agents managed to reliably defeat the prior two competition winners on 
the largest map. \mapname{(4)BloodBath.scmB} is a challenging map for DRL because game 
lengths are significantly longer than on smaller
maps. \bcPPOAgent\ averaged 3,500 steps per game on
\mapname{(4)BloodBath.scmB} compared to around 925 steps on the next largest map,
\mapname{BWDistantResources32x32}. DRL must learn to propagate rewards over longer time
periods, and the observation-action space is so large that DRL can only hope to explore a
fraction of it. A rushing strategy of sending attack units and surplus workers towards
the enemy as soon as possible is a strong strategy on all but the largest map.
\mapname{(4)BloodBath.scmB} seems to reward a build up of forces before attacking, which
DRL struggled to learn.

We hoped imitation learning would mitigate these issues by providing a model that
generates non-zero win-loss rewards and reasonable observation-action pairs.
However, during PPO fine-tuning, a training policy that initially won 40-50\% of
training games, dropped to 20\% midway through training. It eventually recovers to
winning 40\% of training games; however, the fine-tuned policy had a worse evaluation
win rate than the initial supervised policy. This training curve differs from the
smaller map fine-tuning where the training policy quickly won 60\% of
training games and improved upon the evaluation win rate. When transitioning to larger 
maps, \cite{Ferdinand2021doublecone} used a teacher KL-divergence loss term to keep the 
policy from diverging too far from the prior trained policy. Using a similar teacher 
loss term when fine-tuning the behavior cloned policy could help the policy maintain its 
performance.

Extending the curriculum to include more game states could also improve large map
training. For example, varied existing agents can be used to advance the game before
switching to a training agent to finish the game \cite{DBLP:conf/icml/UchenduX0ZYSBFM23}.
Another way is to utilize a prioritized fictitious self-play mechanism used by
\cite{Vinyals2019GrandmasterLI} that prioritizes training the agent on the
most difficult prior agent checkpoints.

While the squnet neural network architectures theoretically have a large enough
receptive field to handle the largest map, using a smaller-map trained model as a
starting point for transfer learning to the largest maps was not effective. One possible
explanation is that the model learns behaviors on a smaller size scale and is not able
to generalize behaviors such as resource gathering or enemy pursuit to larger maps. If
units were instead processed as entities in a transformer architecture, the agent could
process relationships between entities across the map. AlphaStar \cite{Vinyals2019GrandmasterLI} used a transformer
entity encoder as part of its processing.

\subsection{Academic Competition and Research Considerations}
Training multiple models for \agentName\ took 70 GPU-days. Imitation learning \bcAgent\
trained for 23 GPU-days, and PPO fine-tuning \bcPPOAgent\ took another 49 GPU-days. 
These are significant amounts of compute for a mostly academic competition.

There are several ways to make DRL more feasible in a competition and educational setting:
\begin{inparaenum}[(1)]
    \item focus on smaller maps,
    \item fine-tune pretrained models from DRL or behavior cloning,
    \item transfer an existing model to new maps, or
    \item use a significantly smaller neural network architecture.
\end{inparaenum}
The largest map took 19 GPU-days to train for \agentName, 15 GPU-days for \bcAgent, and
34 GPU-days for \bcPPOAgent. Over two-thirds of training time for \bcPPOAgent\ was spent
training on the largest map to little benefit. Reference
\cite{DBLP:journals/corr/abs-2105-13807} trained an agent for player 1 on a single 16x16
map in 2.5 days. Training \bcAgent\ followed by \bcPPOAgent\ on the 5 Open maps up to
size 16x16 took about 7.5 days.

Fine-tuning and transfer learning were critical to making \agentName\ competitive. Both
took significantly less time than training from randomly initialized weights because the
policy already makes reasonable tactical actions and the critic already makes reasonable
value estimates. We didn't train
\bcPPOAgent\ on specific maps, so fine-tuning there could improve win rates. Behavior
cloning other agents (possibly several agents simultaneously), will bootstrap DRL agents
to effective policies that would be extremely difficult to obtain with naive DRL
training.

\agentName's DoubleCone and \bcPPOAgent's deep squnet are relatively large neural
networks, each at around 5 million parameters. Reference
\cite{DBLP:journals/corr/abs-2105-13807}'s best performing policies each used fewer
than 1 million parameters. These smaller networks are quicker to train and have faster
inference time, and there's no definitive evidence so far that they are worse than
larger networks. Reference 
\cite{Winter2023} uses an entity-based transformer exclusively and demonstrates
it can train an agent to beat CoacAI on a 10x10 map in 20 minutes using a single
consumer-grade GPU.

\section{Conclusion}
\agentName\ is the first DRL agent to win the microRTS competition in 6 iterations. It demonstrates that
an iterative training process of fine-tuning and transfer learning is effective for
creating competitive DRL agents. Such a training process can be used by
resource-constrained researchers and students to create novel DRL agents for future
competitions and experiments. Fine-tuning behavior cloning with PPO is a promising way
to create competitive DRL agents without needing to handcraft shaped reward functions.

\ifcogfinal
    \section*{Acknowledgments}
    \addcontentsline{toc}{section}{Acknowledgments}
    We thank the IEEE-CoG 2023 microRTS competition organizers \CompetitionOrganizers\ for
    their work running the competition, helping us work through performance concerns in the
    lead-up to our submission, and providing suggestions for this paper. We thank \ChangLiu\
    for reading and providing feedback on drafts of this paper.
\else
\fi

\bibliographystyle{IEEEtran}
\bibliography{IEEEabrv, cog2024}

\appendices
\section{Competition details}

To participate in the competition, \agentName\ has a Java class that handles turn
handling and resetting commands from the Java game engine. While earlier Python
solutions  passed JSON or XML data over a
socket\footnote{\url{https://github.com/douglasrizzo/python-microRTS}}, \agentName\
passes binary data over a pipe to the Python process as a performance optimization for
the larger maps.

Each agent played every other agent on each map 20 times (10 each as player 1 and 2).
Timeouts were disabled for the competition, but the Java-side of \agentName\ would skip
its turn (submitting no actions) if 100 ms had elapsed. On
\mapname{BWDistantResources32x32}, \agentName\ chose between the DoubleCone and squnet
fine-tuned models by running both models on the first observation 100 times each and
choosing DoubleCone if it computed actions within 75 ms on average.

\begin{table}[t]
    \caption{Policy networks used by \agentName}
    \label{tab:policy-networks}
    \begin{center}
        \begin{tabular}{p{0.5\linewidth}p{0.5\linewidth}}
            \multicolumn{1}{c}{\bf Network}  &\multicolumn{1}{c}{\bf Usage} \\
            \midrule
            ppo-Microrts-finetuned-NoWhereToRun-S1-best & NoWhereToRun9x8 \\ \hline
            ppo-Microrts-A6000-finetuned-coac-mayari-S1-best & All other maps of size 16x16 and smaller \\ \hline
            ppo-Microrts-finetuned-DoubleGame-shaped-S1-best & DoubleGame24x24 \\ \hline
            ppo-Microrts-finetuned-DistantResources-shaped-S1-best & BWDistantResources32x32 if completion time under 75 ms \\ \hline
            ppo-Microrts-squnet-DistantResources-128ch-finetuned-S1-best & BWDistantResources32x32 if completion time above 75 ms \\ \hline
            ppo-Microrts-squnet-map32-128ch-selfplay-S1-best & All other maps where longest dimension is between 17-32 \\ \hline
            ppo-Microrts-squnet-map64-64ch-selfplay-S1-best & Maps where the longest
            dimension is over 32 \\
        \end{tabular}
    \end{center}
\end{table}

\clearpage
\onecolumn

\section{Neural network architecture}
DoubleCone(4, 6, 4) \cite{Ferdinand2021doublecone} consists of
\begin{inparaenum}[(1)]
    \item 4 residual blocks;
    \item a downscaled residual block consisting of a stride-4 convolution, 6 residual blocks, and
        2 stride-2 transpose convolutions;
    \item 4 residual blocks; and
    \item actor and value heads (Supplemental Fig.~\ref{suppfig:doublecone}).
\end{inparaenum}
Each residual block includes a squeeze-excitation layer after the second convolutional
layer (Supplemental Fig.~\ref{fig:squeezeexcitation}). The values heads are each 
\begin{inparaenum}[(1)]
    \item 2 stride-2 convolutions,
    \item an adaptive average pooling layer,
    \item flattened,
    \item 2 densely connected layers, and
    \item an activation function (Identity [no activation] or Tanh) to a single, scalar
    value (Supplemental Fig.~\ref{fig:valueheads}).
\end{inparaenum}
The adaptive average pooling layer allows the network to be used on different map sizes.

\begin{suppfigure}[H]
    \begin{center}
        \includegraphics[width=0.4\linewidth]{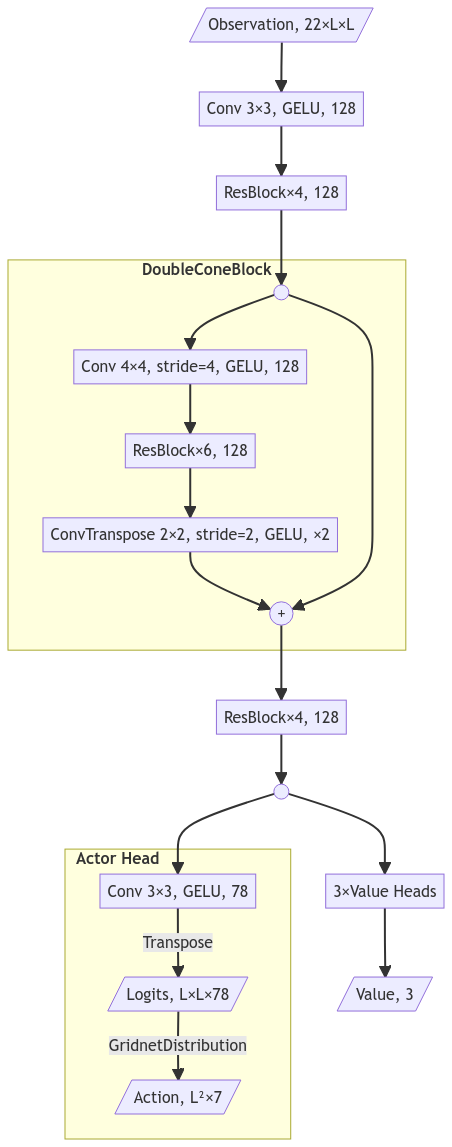}
    \end{center}
    \caption{DoubleCone(4, 6, 4) neural network architecture.}
    \label{suppfig:doublecone}
\end{suppfigure}

\begin{suppfigure}[H]
    \begin{center}
        \includegraphics[width=0.3\linewidth]{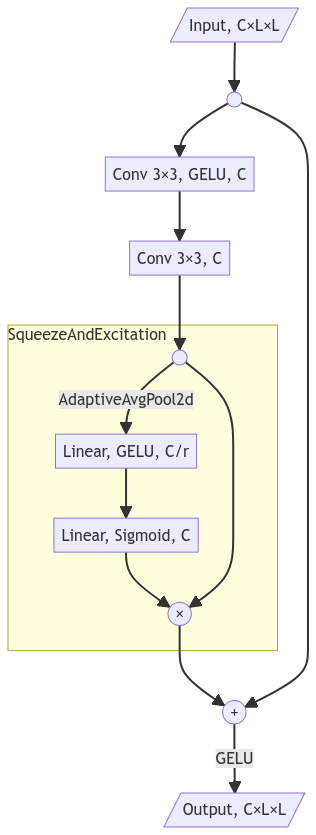}
    \end{center}
    \caption{ResBlock used in DoubleCone, squnet32, and squnet64. The residual block is similar to a standard residual block but inserts a Squeeze-Excitation block after the convolutional layers and before the residual connection.}
    \label{fig:squeezeexcitation}
\end{suppfigure}

\begin{suppfigure}[H]
    \begin{center}
        \includegraphics[width=0.7\linewidth]{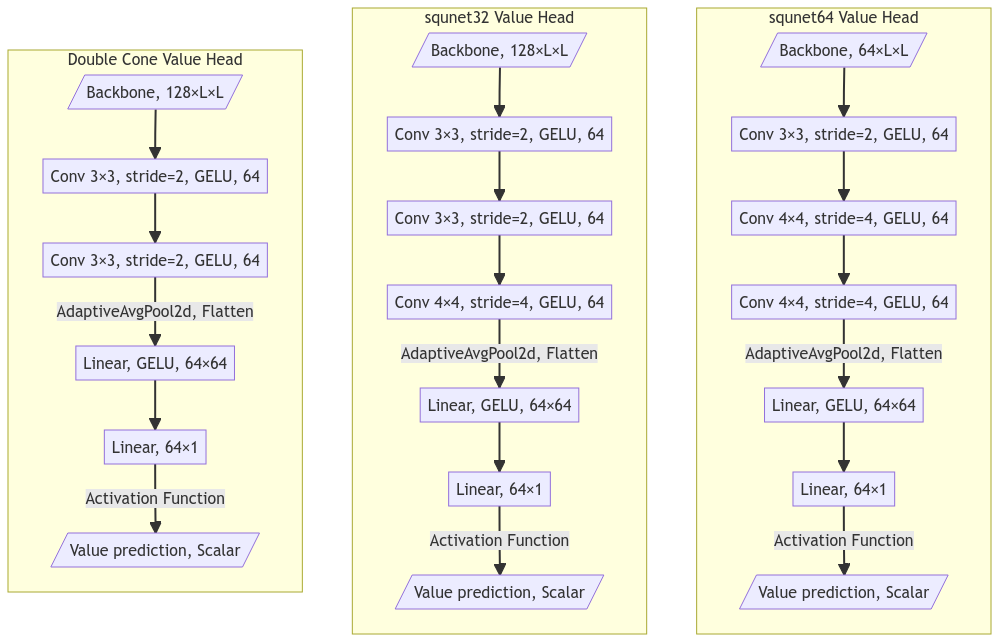}
    \end{center}
    \caption{Value heads used in (from left to right) DoubleCone, squnet32, and
    squnet64. The AdaptiveAvgPool2d layer allows the network to be used on various map
    sizes.}
    \label{fig:valueheads}
\end{suppfigure}

\begin{suppfigure}[H]
    \begin{center}
        \includegraphics[width=0.45\linewidth]{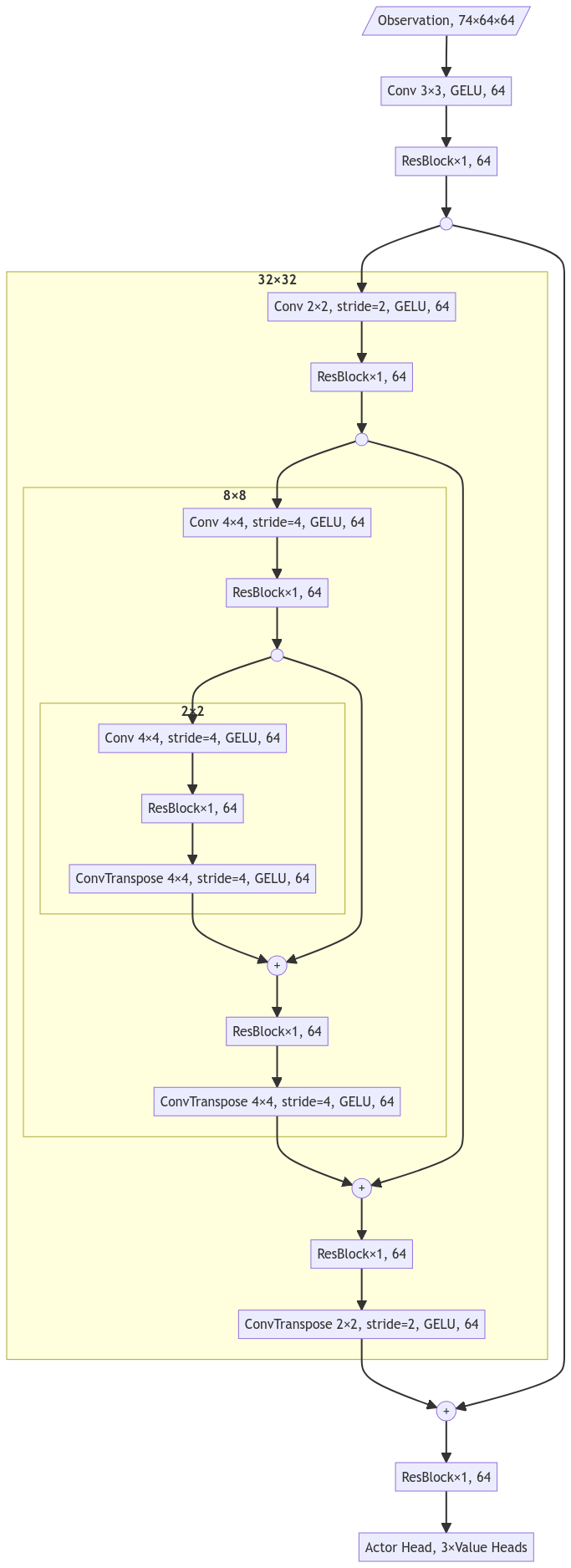}
    \end{center}
    \caption{squnet64 neural network architecture. Instead of one downscaling block as in DoubleCone, this network downscales 3 times. This aggressive downscaling reduces the number of computations for larger maps, while theoretically supporting a large receptive field.}
    \label{fig:squnet}
\end{suppfigure}
        
\begin{supptable}[h]
    \centering
    \begin{threeparttable}
        \caption{Comparison of different architectures}
        \label{table:architectureBreakdown}
        \begin{tabular}{cccc}
            & \textbf{DoubleCone} & \textbf{squnet-map32\tnote{a}} & \textbf{squnet-map64} \\
            \midrule
            Levels & 2 & 4 & 4 \\
            Encoder residual blocks/level & [4, 6] & [1, 1, 1, 1] & [1, 1, 1, 1] \\
            Decoder residual blocks/level & [4] & [1, 1, 1] & [1, 1, 1] \\
            Stride/level & [4] & [2, 2, 4] & [2, 4, 4] \\
            Deconvolution strides/level & [[2, 2]\tnote{b}{ }] & [2, 2, 4] & [2, 4, 4] \\
            Channels/level & [128, 128] & [128, 128, 128, 128] & [64, 64, 64, 64] \\
            Trainable parameters & 5,014,865 & 3,584,657 & 1,420,625 \\
            MACs\tnote{c} & \begin{tabular}[c]{@{}c@{}}0.70B (16x16)\tnote{d} \\ 0.40B (12x12)\tnote{e} \\ 1.58B (24x24) \\ 2.81B (32x32)\end{tabular} & 1.16B (32x32) & 1.41B (64x64) \\ 
        \end{tabular}
        \begin{tablenotes}
            \item[a] Used by ppo-Microrts-squnet-DistantResources-128ch-finetuned-S1-best and ppo-Microrts-squnet-map32-128ch-selfplay-S1-best. 
            \item[b] 2 stride-2 transpose convolutions to match the 1 stride-4 convolution.
            \item[c] Multiply-Accumulates for computing actions for a single observation.
            \item[d] All maps smaller than 16x16 (except NoWhereToRun9x8) are padded with walls up to 16x16. 
            \item[e] NoWhereToRun9x8 is padded with walls up to 12x12.
        \end{tablenotes}
    \end{threeparttable}
\end{supptable}

\section{Initial training details}
\label{appendix:initial-training-details}
\agentName\ was trained with partial observability and environment non-determinism disabled.

\begin{supptable}[H]
    \centering
    \begin{threeparttable}
    \caption{Initial training schedule from a randomly initialized model}
    \label{tab:initial-training-schedule}
    \begin{tabular}{lccccr}
    \toprule
     & Phase 1 & Transition 1→2\tnote{a} & Phase 2 & Transition 2→3\tnote{a} & Phase 3 \\
    \midrule
    Steps & 90M & 60M & 30M & 60M & 60M \\
    Reward Weights\tnote{b} & [0.8, 0.01, 0.19] &  & [0, 0.5, 0.5] &  & [0, 0.99, 0.01] \\
    $c_1$ (Value Loss Coef)\tnote{b} & [0.5, 0.1, 0.2] &  & [0, 0.4, 0.4] &  & [0, 0.5, 0.1]\\
    $c_2$ (Entropy Coef)& 0.01 & & 0.01 & & 0.001 \\
    Learning Rate & $10^{-4}$ & & $10^{-4}$ & & $5 \times 10^{-5}$\\
    \bottomrule
    \end{tabular}
    \begin{tablenotes}
       \item[a] Values are linearly interpolated between phases based on step count.
       \item[b] Listed weights are for the shaped, win-loss, cost-based values, respectively.
    \end{tablenotes}
    \end{threeparttable}
\end{supptable}

\begin{supptable}[H]
    \centering
    \begin{threeparttable}
    \caption{Comparison of initial training, shaped fine-tuning, and sparse fine-tuning parameters}
    \label{tab:training-parameters}
    \begin{tabular}{lccc}
    Parameter & Initial Training & Shaped Fine-Tuning & Sparse Fine-Tuning \\
    \midrule
    Steps & 300M &   100M &   100M \\
    Number of Environments & 24 &    \textquotedbl &    \textquotedbl \\
    Rollout Steps Per Env & 512 &   \textquotedbl &   \textquotedbl \\
    Minibatch Size & 4096 &   \textquotedbl &   \textquotedbl \\
    Epochs Per Rollout & 2 &   \textquotedbl &   \textquotedbl \\
    $\gamma$ (Discount Factor) & [0.99, 0.999, 0.999]\tnote{a} &   \textquotedbl &   \textquotedbl \\
    GAE $\lambda$ & [0.95, 0.99, 0.99]\tnote{a} &   \textquotedbl &   \textquotedbl \\
    Clip Range & 0.1 &   \textquotedbl &   \textquotedbl \\
    Clip Range VF & 0.1 &    \textquotedbl &   \textquotedbl \\
    VF Coef Halving\tnote{b} & True  &    \textquotedbl &   \textquotedbl \\
    Max Grad Norm &  0.5 &   \textquotedbl &   \textquotedbl \\   
    Latest Self-play Envs   &   12 &                      \textquotedbl &                      \textquotedbl \\
    Old Self-play Envs   &   12 &                      0 &                      0 \\
    Bots   &   none & CoacAI: 12 & \begin{tabular}[c]{@{}c}CoacAI: 6\\ Mayari: 6\end{tabular} \\
    Maps   &   \begin{tabular}[c]{@{}c}basesWorkers16x16A \\ TwoBasesBarracks16x16 \\
    basesWorkers8x8A \\ FourBasesWorkers8x8 \\ NoWhereToRun9x8 \\
    EightBasesWorkers16x16\tnote{c} \end{tabular} &  \textquotedbl & \textquotedbl \\
    \end{tabular}
    \begin{tablenotes}
    \item[\textquotedbl] Same value as cell to left.
    \item[a] Value per value head (shaped, win-loss, cost-based).
    \item[b] Multiply $v_{\text{loss}}$ by 0.5, as done in CleanRL.
    \item[c] Map not used in competition.
    \end{tablenotes}
    \end{threeparttable}
\end{supptable}

\begin{supptable}[H]
    \centering
    \begin{threeparttable}
    \caption{Shaped fine-tuning schedule}
    \label{tab:shaped-finetuning-schedule}
    \begin{tabular}{lccccc}
    \toprule
     & Start & Transition →1\tnote{a} & Phase 1 & Transition 1→2\tnote{a} & Phase 2 \\
     \midrule
    Steps & & 5M & 30M & 20M & 45M \\
    Reward Weights\tnote{b} & [0, 0.99, 0.01] & & [0, 0.5, 0.5] & & [0, 0.99, 0.01] \\
    $c_1$ (Value Loss Coef)\tnote{b} & [0, 0.4, 0.2] & & [0, 0.4, 0.4] & & [0, 0.5, 0.1]\\
    $c_2$ (Entropy Coef) & 0.01 & & 0.01 & & 0.001 \\
    Learning Rate & $10^{-5}$ & & $5 \times 10^{-5}$ & & $5 \times 10^{-5}$\\
    \bottomrule
    \end{tabular}
    \begin{tablenotes}
       \item[a] Values are linearly interpolated between phases based on step count.
       \item[b] Listed weights are for the shaped, win-loss, cost-based values, respectively.
    \end{tablenotes}
    \end{threeparttable}
\end{supptable}

\begin{supptable}[H]
    \centering
    \begin{threeparttable}
    \caption{Sparse fine-tuning schedule}
    \label{tab:sparse-finetuning-schedule}
    \begin{tabular}{lcccc}
    \toprule
     & Phase 1 & Transition 1→2\tnote{a} & Phase 2 \\
     \midrule
    Steps & 30M & 40M & 30M \\
    Reward Weights\tnote{b} & [0, 0.99, 0.01] &  & [0, 0.99, 0.01] \\
    $c_1$ (Value Loss Coef)\tnote{b} & [0, 0.5, 0.1] &  & [0, 0.5, 0.1]\\
    $c_2$ (Entropy Coef) & 0.001 & & 0.0001 \\
    Learning Rate & $5 \times 10^{-5}$ & & $10^{-5}$ \\
    \bottomrule
    \end{tabular}
    \begin{tablenotes}
       \item[a] Values are linearly interpolated between phases based on step count.
       \item[b] Listed weights are for the shaped, win-loss, cost-based values, respectively.
    \end{tablenotes}
    \end{threeparttable}
\end{supptable}

\section{Transfer learning details}
\label{appendix:transfer-learning-details}

\begin{supptable}[H]
    \centering
    \begin{threeparttable}
    \caption{Transfer learning schedule starting from ppo-Microrts-A6000-finetuned-coac-mayari-S1-best model}
    \label{tab:transfer-learning-schedule}
    \begin{tabular}{lccccc}
    \toprule
     & Start & Transition →1\tnote{a} & Phase 1 & Transition 1→2\tnote{a} & Phase 2 \\
     \midrule
    Steps & & 5M & 30M & 20M & 45M \\
    Reward Weights\tnote{b} & [0, 0.99, 0.01] & & [0.4, 0.5, 0.1] & & [0, 0.99, 0.01] \\
    $c_1$ (Value Loss Coef)\tnote{b} & [0.2, 0.4, 0.2] & & [0.3, 0.4, 0.1] & & [0, 0.5, 0.1]\\
    $c_2$ (Entropy Coef) & 0.01 & & 0.01 & & 0.0001 \\
    Learning Rate & $5 \times 10^{-5}$ & & $7 \times 10^{-5}$ & & $10^{-5}$ \\
    \bottomrule
    \end{tabular}
    \begin{tablenotes}
       \item[a] Values are linearly interpolated between phases based on step count.
       \item[b] Listed weights are for the shaped, win-loss, cost-based values, respectively.
    \end{tablenotes}
    \end{threeparttable}
\end{supptable}

\section{squnet learning details}
\label{appendix:squnet-learning-details}

\begin{supptable}[H]
    \centering
    \begin{threeparttable}
    \caption{Squnet training parameters}
    \label{tab:squnet-training-parameters}
    \begin{tabular}{lccc}
    \toprule
    Parameter & map32 & map32-DistantResources & map64 \\
    \midrule
    Steps & 200M & 100M & 200M \\
    Number of Environments & 24 & \textquotedbl & \textquotedbl \\
    Rollout Steps Per Env & 512 & 512 & 256 \\
    Minibatch Size & 2048 & 2048 & 258 \\
    Clip Range & 0.1 & \textquotedbl & \textquotedbl \\
    Clip Range VF & none & \textquotedbl & \textquotedbl \\
    Latest Self-play Envs & 12 & \textquotedbl & \textquotedbl \\
    Old Self-play Envs & 6 & 6 & 4 \\
    Bots & \begin{tabular}[c]{@{}c}CoacAI: 3 \\ Mayari: 3 \end{tabular} & \begin{tabular}[c]{@{}c}CoacAI: 3 \\ Mayari: 3 \end{tabular} & \begin{tabular}[c]{@{}c}CoacAI: 4 \\ Mayari: 4 \end{tabular} \\
    Maps & \begin{tabular}[c]{@{}c}DoubleGame24x24 \\ BWDistantResources32x32 \\ chambers32x32\tnote{a} \end{tabular} & BWDistantResources32x32 & \begin{tabular}[c]{@{}c}BloodBath.scmB \\ BloodBath.scmE\tnote{a}\end{tabular} \\   
    \bottomrule
    \end{tabular}
    \begin{tablenotes}
    \item[\textquotedbl] Same value as cell to left.
    \item[a] Not competition Open maps.
    \end{tablenotes}
    \end{threeparttable}
\end{supptable}

\begin{supptable}[H]
    \centering
    \begin{threeparttable}
    \caption{squnet training schedule starting with randomly initialized weights}
    \label{tab:squnet-training-schedule}
    \begin{tabular}{lcccc}
    \toprule
     & Phase 1 & Transition 1→2\tnote{a} & Phase 2 \\
     \midrule
    Steps & 100M & 60M & 40M \\
    Reward Weights\tnote{b} & [0.8, 0.01, 0.19] &  & [0, 0.99, 0.01] \\
    $c_1$ (Value Loss Coef)\tnote{b} & [0.5, 0.1, 0.2] &  & [0, 0.5, 0.1]\\
    $c_2$ (Entropy Coef) & 0.01 & & 0.001 \\
    Learning Rate & $10^{-4}$ & & $5 \times 10^{-5}$ \\
    \bottomrule
    \end{tabular}
    \begin{tablenotes}
       \item[a] Values are linearly interpolated between phases based on step count.
       \item[b] Listed weights are for the shaped, win-loss, cost-based values, respectively.
    \end{tablenotes}
    \end{threeparttable}
\end{supptable}

\section{Behavior cloning details}
\label{appendix:behavior-cloning-details}
\begin{supptable}[H]
    \centering
    \begin{threeparttable}
    \caption{Neural architecture for behavior cloning and PPO fine-tuned training}
    \label{tab:bc-architecture}
    \begin{tabular}{lc}
    \toprule
                                 & deep16-128 \\
    \midrule
    Levels                      & 3  \\
    Encoder residual blocks/level & [3, 2, 4] \\
    Decoder residual blocks/level & [3, 2] \\
    Stride per level            & [4, 4] \\
    Deconvolution strides per level & [[2, 2], [2, 2]]\tnote{a}\\
    Channels per level          & [128, 128, 128] \\
    Trainable parameters        & 5,027,279 \\
    MACs\tnote{b} (16x16)          & 0.52B \\
    MACs\tnote{b} (64x64)          & 8.40B \\
    \bottomrule
    \end{tabular}
    \begin{tablenotes}
       \item[a] 2 stride-2 transpose convolutions to match each 1 stride-4 convolution.
       \item[b] Multiply-Accumulates for computing actions for a single observation.
    \end{tablenotes}
    \end{threeparttable}
\end{supptable}

\begin{supptable}[H]
    \centering
    \begin{threeparttable}
    \caption{Behavior cloning training parameters}
    \label{tab:bc-training-parameters}
    \begin{tabular}{lccc}
    Map Size & 16x16 & 32x32 & 64x64 \\
    \midrule
    Steps & 100M & \textquotedbl & \textquotedbl \\
    Number of Environments & 36 & 24 & 24 \\
    Rollout Steps Per Env & 512 & \textquotedbl & \textquotedbl \\
    Minibatch Size & 3072 & 768 & 192 \\
    Epochs Per Rollout & 2 & \textquotedbl & \textquotedbl \\
    $\gamma$ (Discount Factor) & 0.999 & 0.9996 & 0.999 \\
    GAE $\lambda$ & 0.99 & 0.996 & 0.999 \\
    Max Grad Norm & 0.5 & \textquotedbl & \textquotedbl \\
    Gradient Accumulation & FALSE & FALSE & TRUE \\
    Scale Loss by \# Actions & TRUE & \textquotedbl & \textquotedbl \\
    \hline
    Bots & \begin{tabular}[c]{@{}c@{}}Mayari: 12\\ CoacAI: 12\\ POLightRush:
    12\end{tabular} & \begin{tabular}[c]{@{}c@{}}Mayari: 12\\ CoacAI: 6\\ POLightRush:
    6\end{tabular} & \begin{tabular}[c]{@{}c@{}}Mayari: 8\\ CoacAI: 8\\ POLightRush:
    8\end{tabular} \\
    \hline
    Maps & \begin{tabular}[c]{@{}c@{}}basesWorkers16x16A\\ TwoBasesBarracks16x16\\
    basesWorkers8x8A\\ FourBasesWorkers8x8\\ NoWhereToRun9x8\\
    EightBasesWorkers16x16\end{tabular} & \begin{tabular}[c]{@{}c@{}}DoubleGame24x24\\
    BWDistantResources32x32\\ chambers32x32\end{tabular} &
    \begin{tabular}[c]{@{}c@{}}(4)BloodBath.scmB\\ (4)BloodBath.scmE\end{tabular}
    \end{tabular}
    \begin{tablenotes}
    \item[\textquotedbl] Same value as cell to left.
    \end{tablenotes}
\end{threeparttable}
\end{supptable}

\begin{supptable}[H]
    \centering
    \begin{threeparttable}
    \caption{Training parameters for PPO fine-tuning of behavior cloned models}
    \label{tab:bc-ppo-training-parameters}
    \begin{tabular}{lccc}
    Map Size & 16x16 & 32x32 & 64x64 \\
    \midrule
    Steps & 100M & 200M & 200M \\
    Number of Environments & 36 & 24 & 48 \\
    Rollout Steps Per Env & 512 & \textquotedbl & \textquotedbl \\
    Minibatch Size & 3072 & 768 & 192 \\
    Epochs Per Rollout & 2 & \textquotedbl & \textquotedbl \\
    {\color[HTML]{1A1A1A} $\gamma$ (Discount Factor)} & 0.999 & 0.9996 & 0.99983 \\
    GAE $\lambda$ & 0.99 & 0.996 & 0.9983 \\
    Clip Range & 0.1 & \textquotedbl & \textquotedbl \\
    Clip Range VF & none & \textquotedbl & \textquotedbl \\
    VF Coef Halving\tnote{a} & TRUE & \textquotedbl & \textquotedbl \\
    Max Grad Norm & 0.5 & \textquotedbl & \textquotedbl \\
    Gradient Accumulation & FALSE & TRUE & TRUE \\
    Latest Selfplay Envs & 12 & 12 & 28 \\
    Old Selfplay Envs & 12 & 6 & 12 \\
    Bots & \begin{tabular}[c]{@{}c@{}}Mayari: 6\\ CoacAI: 6\end{tabular} & \begin{tabular}[c]{@{}c@{}}Mayari: 3\\ CoacAI: 3\end{tabular} & \begin{tabular}[c]{@{}c@{}}Mayari: 2\\ CoacAI: 2\\ POLightRush: 2\\ POWorkerRush: 2\end{tabular} \\
    Maps & \begin{tabular}[c]{@{}c@{}}basesWorkers16x16A\\ TwoBasesBarracks16x16\\ basesWorkers8x8A\\ FourBasesWorkers8x8\\ NoWhereToRun9x8\\ EightBasesWorkers16x16\end{tabular} & \begin{tabular}[c]{@{}c@{}}DoubleGame24x24\\ BWDistantResources32x32\\ chambers32x32\end{tabular} & \begin{tabular}[c]{@{}c@{}}(4)BloodBath.scmB\\ (4)BloodBath.scmE\end{tabular}
    \end{tabular}
    \begin{tablenotes}
    \item[\textquotedbl] Same value as cell to left.
    \item[a] Multiply $v_{\text{loss}}$ by 0.5, as done in CleanRL.
    \end{tablenotes}
\end{threeparttable}
\end{supptable}

\begin{supptable}[H]
    \caption{Behavior cloning schedule for 16x16 maps. Values in transition are linearly interpolated.}
    \label{tab:bc-schedule-map16}
    \begin{center}
    \begin{tabular}{lccc}
    & Start & Transition & End \\
    \midrule
    & & 100M & \\
    Learning Rate & $8 \times 10^{-5}$ &  & 0 \\
    \end{tabular}
\end{center}
\end{supptable}

\begin{supptable}[H]
    \caption{Behavior cloning schedule for 32x32 and 64x64 maps. Values in transitions are cosine interpolated.}
    \label{tab:bc-schedule}
    \begin{center}
    \begin{tabular}{lccccc}
    & Start & Transition →1 & Phase 1 & Transition 1→2 & Phase 2 \\
    \midrule
    &  & 5M & 5M & 85M & 5M \\
    Learning Rate & $10^{-5}$ &  & $8 \times 10^{-5}$ &  & $10^{-6}$
    \end{tabular}
\end{center}
\end{supptable}

\begin{supptable}[H]
    \caption{Schedule for PPO fine-tuning of behavior cloned model for 16x16 map. Transition values are cosine interpolated.}
    \label{tab:bc-ppo-schedule-map16}
    \begin{center}
    \begin{tabular}{lccccc}
    & Start & Transition →1 & Phase 1 & Transition 1→2 & Phase 2 \\
    \midrule
    &  & 5M & 5M & 85M & 5M \\
    $c_2$ (Entropy Coef) & 0.001 &  & 0.001 &  & 0.0001 \\
    Learning Rate & $10^{-5}$ &  & $5 \times 10^{-5}$ &  & $10^{-5}$ \\
    \end{tabular}
\end{center}
\end{supptable}

\begin{supptable}[H]
    \caption{Schedule for PPO fine-tuning of behavior cloned model for 32x32 map. Transition values are cosine interpolated.}
    \label{tab:bc-ppo-schedule-map32}
    \begin{center}
    \begin{tabular}{lccccc}
    & Start & Transition →1 & Phase 1 & Transition 1→2 & Phase 2 \\
    \midrule
    &  & 10M & 80M & 70M & 40M \\
    $c_2$ (Entropy Coef) & 0.001 &  & 0.001 &  & 0.0001 \\
    Learning Rate & $10^{-5}$ &  & $5 \times 10^{-5}$ &  & $10^{-5}$ \\
    \end{tabular}
\end{center}
\end{supptable}

\begin{supptable}[H]
    \caption{Schedule for PPO fine-tuning of behavior cloned model for 64x64 map. Transition values are cosine interpolated. Transition 1→2 being empty means values jump from Phase 1 to Phase 2.}
    \label{tab:bc-ppo-schedule-map64}
    \begin{center}
        \begin{tabular}{lccccccccc}
            & \begin{sideways} Start \end{sideways} & \begin{sideways} Transition →1
            \end{sideways} & \begin{sideways} Phase 1 \end{sideways} & \begin{sideways}
            Transition 1→2 \end{sideways} & \begin{sideways} Phase 2 \end{sideways} &
            \begin{sideways} Transition 2→3 \end{sideways} & \begin{sideways} Phase 3
            \end{sideways} & \begin{sideways} Transition 3→4 \end{sideways} &
            \begin{sideways} Phase 4 \end{sideways} \\
            \midrule
            &  & 10M &  &  &  & 40M & 80M & 66M & 4M \\
           $c_2$ (Entropy Coef) & 0 &  & 0 &  & 0.001 &  & 0.001 &  & 0.0001 \\
           Learning Rate & $10^{-6}$ &  & $5 \times 10^{-5}$ & & $10^{-6}$ &  & $5 \times 10^{-5}$ &  & $10^{-6}$ \\
           \begin{tabular}[c]{@{}l@{}}Freeze Backbone \\ and Policy Head\end{tabular} & TRUE &  & TRUE &  & FALSE &  & FALSE &  & FALSE
    \end{tabular}
\end{center}
\end{supptable}

\section{Training durations}
We trained using Lambda Labs GPU on-demand instances. We used single Nvidia GPU instances, but
different GPUs to be able to fit larger minibatches onto the GPU. A10 (24 GB VRAM) and
A100 (40 GB VRAM) machines had 30 vCPUs and 200 GB RAM. A6000 (48 GB VRAM) machines had
14 vCPUs and 100 GB RAM. We did not fully utilize the CPU, RAM, GPU compute, or hard drive resources
during training.

Behavior cloning and PPO fine-tuning of behavior cloned models were trained only using
A10 machines. We had implemented gradient accumulation at this point to support larger
batch sizes that did not need to fit on the GPU all-at-once.

\label{appendix:training-durations}
\begin{supptable}[H]
    \caption{\agentName\ training durations. Blank models are intermediate models that
    lead to the next row. For example, the first 3 runs are intermediate models for
    16x16. Runs are uploaded to the \wbProject, except
    for squnet-DistantResources (\microRTSWBProjectPath).}
    \label{tab:training-durations}
    \begin{center}
    \begin{tabular}{lccc}
        \multicolumn{1}{c}{Map} & Run ID & GPU & Days Training \\
        \midrule
         & \href{https://wandb.ai/\wbProjectPath/runs/df4flrs4}{\texttt{df4flrs4}} & A10 & 12.5 \\
         & \href{https://wandb.ai/\wbProjectPath/runs/9bz7wsuv}{\texttt{9bz7wsuv}} & A6000 & 2.7 \\
         & \href{https://wandb.ai/\wbProjectPath/runs/tff7xk4b}{\texttt{tff7xk4b}} & A6000 & 4.1 \\
        \multicolumn{1}{c}{16x16} & \href{https://wandb.ai/\wbProjectPath/runs/1ilo9yae}{\texttt{1ilo9yae}} & A6000 & 4.3 \\
         \hline
         & \href{https://wandb.ai/\wbProjectPath/runs/hpp5pffx}{\texttt{hpp5pffx}} & A10 & 1.9 \\
        \multicolumn{1}{c}{NoWhereToRun9x8} & \href{https://wandb.ai/\wbProjectPath/runs/vmns9sbe}{\texttt{vmns9sbe}} & A10 & 1.7 \\
         \hline
        \multicolumn{1}{c}{DoubleGame24x24} & \href{https://wandb.ai/\wbProjectPath/runs/unnxtprk}{\texttt{unnxtprk}} & A6000 & 5.3 \\
         \hline
        \multicolumn{1}{c}{BWDistantResources32x32} & \href{https://wandb.ai/\wbProjectPath/runs/x4tg80vk}{\texttt{x4tg80vk}} & A100 & 3.6 \\
         \hline
        \multicolumn{1}{c}{32x32} & \href{https://wandb.ai/\wbProjectPath/runs/tga53t25}{\texttt{tga53t25}} & A6000 & 10.2 \\
        \multicolumn{1}{c}{squnet-DistantResources} & \href{https://wandb.ai/\microRTSWBProjectPath/runs/jl8zkpfr}{\texttt{jl8zkpfr}} & A6000 & 5.0 \\
         \hline
        \multicolumn{1}{c}{64x64} & \href{https://wandb.ai/\wbProjectPath/runs/nh5pdv4o}{\texttt{nh5pdv4o}} & A6000 & 19.0 \\
        \hline
         & \multicolumn{1}{l}{} & \multicolumn{1}{l}{} & 70.4 \\
         & \multicolumn{1}{l}{} & \multicolumn{1}{l}{} & \multicolumn{1}{l}{}
    \end{tabular}
    \end{center}
\end{supptable}

\begin{supptable}[H]
    \caption{Behavior cloning training durations. Runs are uploaded to the \microRTSWBProject.}
    \label{tab:bc-training-durations}
    \begin{center}
    \begin{tabular}{llc}
        Map Size & Run ID & Days Training \\
        \midrule
        16x16 & \href{https://wandb.ai/\microRTSWBProjectPath/runs/lhs1b2gj}{\texttt{lhs1b2gj}} & 3.5 \\
        32x32 & \href{https://wandb.ai/\microRTSWBProjectPath/runs/16o4391r}{\texttt{16o4391r}} & 4.7 \\
        64x64 & \href{https://wandb.ai/\microRTSWBProjectPath/runs/uksp6znl}{\texttt{uksp6znl}} & 15.1 \\
        \hline
        \multicolumn{1}{l}{} & \multicolumn{1}{l}{} & 23.3
    \end{tabular}
    \end{center}
\end{supptable}

\begin{supptable}[H]
    \caption{Training durations for PPO fine-tuning of behavior cloned models. Runs are uploaded to the \microRTSWBProject.}
    \label{tab:bc-ppo-training-durations}
    \begin{center}
    \begin{tabular}{llc}
        Map Size & Run ID & Days Training \\
        \midrule
        16x16 & \href{https://wandb.ai/\microRTSWBProjectPath/runs/a4efzeug}{\texttt{a4efzeug}} & 4.0 \\
        32x32 & \href{https://wandb.ai/\microRTSWBProjectPath/runs/042rwd8p}{\texttt{042rwd8p}} & 11.3 \\
        64x64 & \href{https://wandb.ai/\microRTSWBProjectPath/runs/9l2debnz}{\texttt{9l2debnz}} & 33.9 \\
        \hline
        \multicolumn{1}{l}{} & \multicolumn{1}{l}{} & 49.1
    \end{tabular}
    \end{center}
\end{supptable}

\section{Single player round-robin benchmark setup}
\label{appendix:single-player-benchmark-setup}
In Section~\ref{sec:single-player-benchmark}, \agentName\ plays on the 8 Open maps
against 4 opponents:
\begin{inparaenum}[(1)]
    \item baseline POWorkerRush,
    \item baseline and 2017 competition winner POLightRush,
    \item 2020 competition winner CoacAI, and
    \item last competition (2021) winner Mayari.
\end{inparaenum}
\agentName\ normally plays against each opponent on each map for 100 matches (50
each as player 1 and 2). The exception is the squnet model for \mapname{BWDistantResources32x32},
where \agentName\ only played each opponent for 20 matches (10 each as player 1 and 2).
All opponents use A* for pathfinding, which is default for competitions.  Win rates are 
percentages of wins where draws count as 0.5 wins for each player. The single player 
round-robin benchmark was run on a 2018 Mac Mini with Intel i7-8700B CPU (6-core, 
3.2GHz) with PyTorch limited to 6 threads. Timeouts were set to 100 ms. If
an agent took 20ms over the deadline (120 ms total), the game was terminated and the win
awarded to the opponent.

In Section~\ref{sec:behavior-cloning-results}, \bcAgent\ and \bcPPOAgent\ play each
opponent on each map for 20 matches
(10 each as player 1 and 2).

\section{Additional IEEE-CoG 2023 microRTS competition details}
\begin{supptable}[H]
    \caption{Win rates of all agents in the IEEE-CoG 2023 microRTS competition on Open maps. Player 1 is the row agent and player 2 is the column agent. Each win rate value is the percentage of games won by player 1 (the row agent). Higher win rates are redder. Lower win rates are bluer.}
    \label{tab:all-competition-winrate}
    \begin{center}
    \arrayrulecolor{black}
    \begin{tabular}{lccccccccccccccccc|c}
    & \begin{sideways} Mayari \end{sideways} & \begin{sideways} 2L \end{sideways} 
    & \begin{sideways} \textbf{\agentName} \end{sideways} & \begin{sideways} ObiBotKenobi \end{sideways} 
    & \begin{sideways} Aggrobot \end{sideways} & \begin{sideways} sophia \end{sideways} 
    & \begin{sideways} bRHEAdBot \end{sideways} & \begin{sideways} Ragnar \end{sideways} 
    & \begin{sideways} POLightRush \end{sideways} & \begin{sideways} SaveTheBeesV4 \end{sideways} 
    & \begin{sideways} POWorkerRush \end{sideways} & \begin{sideways} MyMicroRtsBot \end{sideways} 
    & \begin{sideways} NaiveMCTS \end{sideways} & \begin{sideways} myBot \end{sideways} 
    & \begin{sideways} NIlSiBot \end{sideways} & \begin{sideways} Predator \end{sideways} 
    & \begin{sideways} RandomBiasedAI \end{sideways} & \begin{sideways} Overall \end{sideways} \\
    \arrayrulecolor{black}\specialrule{.5pt}{0pt}{0pt}
    Mayari         & -      & \colcellnobold{53} & \colcellnobold{32} & \colcellnobold{73} & \colcellnobold{78} & \colcellnobold{93} & \colcellnobold{95} & \colcellnobold{64} & \colcellnobold{88} & \colcellnobold{93} & \colcellnobold{75} & \colcellnobold{78} & \colcellnobold{100} & \colcellnobold{100} & \colcellnobold{100} & \colcellnobold{100} & \colcellnobold{100} & \colcellnobold{82} \\
    2L             & \colcellnobold{51} & -  & \colcellnobold{39} & \colcellnobold{50} & \colcellnobold{69} & \colcellnobold{63} & \colcellnobold{93} & \colcellnobold{56} & \colcellnobold{75} & \colcellnobold{98} & \colcellnobold{88} & \colcellnobold{81} & \colcellnobold{76} & \colcellnobold{94} & \colcellnobold{94} & \colcellnobold{95} & \colcellnobold{96} & \colcellnobold{76} \\
    \textbf{\agentName} & \colcellnobold{62} & \colcellnobold{59} & -  & \colcellnobold{49} & \colcellnobold{64} & \colcellnobold{71} & \colcellnobold{64} & \colcellnobold{64} & \colcellnobold{64} & \colcellnobold{78} & \colcellnobold{78} & \colcellnobold{76} & \colcellnobold{84} & \colcellnobold{94} & \colcellnobold{73} & \colcellnobold{87} & \colcellnobold{87} & \colcellnobold{72} \\
    ObiBotKenobi   & \colcellnobold{39} & \colcellnobold{29} & \colcellnobold{47} & -  & \colcellnobold{47} & \colcellnobold{69} & \colcellnobold{60} & \colcellnobold{56} & \colcellnobold{58} & \colcellnobold{83} & \colcellnobold{65} & \colcellnobold{76} & \colcellnobold{72} & \colcellnobold{99} & \colcellnobold{79} & \colcellnobold{85} & \colcellnobold{100} & \colcellnobold{66} \\
    Aggrobot       & \colcellnobold{9}  & \colcellnobold{25} & \colcellnobold{26} & \colcellnobold{60} & -  & \colcellnobold{69} & \colcellnobold{55} & \colcellnobold{44} & \colcellnobold{63} & \colcellnobold{86} & \colcellnobold{69} & \colcellnobold{94} & \colcellnobold{66} & \colcellnobold{94} & \colcellnobold{94} & \colcellnobold{91} & \colcellnobold{94} & \colcellnobold{65} \\
    sophia         & \colcellnobold{25} & \colcellnobold{44} & \colcellnobold{30} & \colcellnobold{35} & \colcellnobold{38} & - & \colcellnobold{41} & \colcellnobold{88} & \colcellnobold{75} & \colcellnobold{76} & \colcellnobold{63} & \colcellnobold{69} & \colcellnobold{71} & \colcellnobold{100} & \colcellnobold{75} & \colcellnobold{84} & \colcellnobold{83} & \colcellnobold{62} \\
    bRHEAdBot      & \colcellnobold{4}  & \colcellnobold{7}  & \colcellnobold{24} & \colcellnobold{44} & \colcellnobold{49} & \colcellnobold{69} & -  & \colcellnobold{51} & \colcellnobold{64} & \colcellnobold{79} & \colcellnobold{59} & \colcellnobold{65} & \colcellnobold{83} & \colcellnobold{99} & \colcellnobold{81} & \colcellnobold{96} & \colcellnobold{98} & \colcellnobold{61} \\
    Ragnar         & \colcellnobold{40} & \colcellnobold{50} & \colcellnobold{32} & \colcellnobold{26} & \colcellnobold{50} & \colcellnobold{13} & \colcellnobold{46} & -  & \colcellnobold{44} & \colcellnobold{71} & \colcellnobold{63} & \colcellnobold{69} & \colcellnobold{73} & \colcellnobold{88} & \colcellnobold{81} & \colcellnobold{73} & \colcellnobold{85} & \colcellnobold{56} \\
    POLightRush    & \colcellnobold{0}  & \colcellnobold{25} & \colcellnobold{29} & \colcellnobold{38} & \colcellnobold{31} & \colcellnobold{44} & \colcellnobold{34} & \colcellnobold{38} & -  & \colcellnobold{71} & \colcellnobold{69} & \colcellnobold{69} & \colcellnobold{73} & \colcellnobold{100} & \colcellnobold{75} & \colcellnobold{91} & \colcellnobold{100} & \colcellnobold{55} \\
    SaveTheBeesV4  & \colcellnobold{14} & \colcellnobold{9}  & \colcellnobold{21} & \colcellnobold{43} & \colcellnobold{31} & \colcellnobold{59} & \colcellnobold{38} & \colcellnobold{47} & \colcellnobold{66} & -  & \colcellnobold{50} & \colcellnobold{57} & \colcellnobold{81} & \colcellnobold{86} & \colcellnobold{85} & \colcellnobold{90} & \colcellnobold{93} & \colcellnobold{54} \\
    POWorkerRush   & \colcellnobold{13} & \colcellnobold{13} & \colcellnobold{21} & \colcellnobold{29} & \colcellnobold{31} & \colcellnobold{44} & \colcellnobold{44} & \colcellnobold{56} & \colcellnobold{38} & \colcellnobold{89} & - & \colcellnobold{75} & \colcellnobold{49} & \colcellnobold{94} & \colcellnobold{81} & \colcellnobold{81} & \colcellnobold{96} & \colcellnobold{53} \\
    MyMicroRtsBot  & \colcellnobold{11} & \colcellnobold{13} & \colcellnobold{15} & \colcellnobold{25} & \colcellnobold{38} & \colcellnobold{56} & \colcellnobold{38} & \colcellnobold{56} & \colcellnobold{38} & \colcellnobold{86} & \colcellnobold{44} & - & \colcellnobold{43} & \colcellnobold{94} & \colcellnobold{69} & \colcellnobold{74} & \colcellnobold{92} & \colcellnobold{49} \\
    NaiveMCTS      & \colcellnobold{0}  & \colcellnobold{11} & \colcellnobold{17} & \colcellnobold{22} & \colcellnobold{34} & \colcellnobold{27} & \colcellnobold{15} & \colcellnobold{26} & \colcellnobold{29} & \colcellnobold{69} & \colcellnobold{56} & \colcellnobold{58} & - & \colcellnobold{92} & \colcellnobold{46} & \colcellnobold{60} & \colcellnobold{84} & \colcellnobold{40} \\
    myBot          & \colcellnobold{1}  & \colcellnobold{6}  & \colcellnobold{21} & \colcellnobold{20} & \colcellnobold{39} & \colcellnobold{48} & \colcellnobold{28} & \colcellnobold{41} & \colcellnobold{43} & \colcellnobold{77} & \colcellnobold{39} & \colcellnobold{40} & \colcellnobold{50} & - & \colcellnobold{55} & \colcellnobold{66} & \colcellnobold{66} & \colcellnobold{40} \\
    NIlSiBot       & \colcellnobold{0}  & \colcellnobold{13} & \colcellnobold{18} & \colcellnobold{18} & \colcellnobold{31} & \colcellnobold{25} & \colcellnobold{13} & \colcellnobold{13} & \colcellnobold{31} & \colcellnobold{63} & \colcellnobold{31} & \colcellnobold{38} & \colcellnobold{51} & \colcellnobold{81} & - & \colcellnobold{58} & \colcellnobold{73} & \colcellnobold{35} \\
    Predator       & \colcellnobold{1}  & \colcellnobold{7}  & \colcellnobold{13} & \colcellnobold{6}  & \colcellnobold{12} & \colcellnobold{21} & \colcellnobold{11} & \colcellnobold{16} & \colcellnobold{14} & \colcellnobold{56} & \colcellnobold{22} & \colcellnobold{28} & \colcellnobold{44} & \colcellnobold{73} & \colcellnobold{43} & - & \colcellnobold{45} & \colcellnobold{26} \\
    RandomBiasedAI & \colcellnobold{0}  & \colcellnobold{1}  & \colcellnobold{15} & \colcellnobold{0}  & \colcellnobold{4}  & \colcellnobold{15} & \colcellnobold{6}  & \colcellnobold{9}  & \colcellnobold{4}  & \colcellnobold{52} & \colcellnobold{4}  & \colcellnobold{13} & \colcellnobold{18} & \colcellnobold{85} & \colcellnobold{39} & \colcellnobold{39} & - & \colcellnobold{19} \\
    \end{tabular}
    \end{center}
\end{supptable}

\section{Additional behavior cloning benchmarks}

\begin{supptable}[H]
    \caption{\bcAgent\ win rate in single player round-robin benchmark. Win rates above 50\% are bolded. Higher win rates are redder. Lower win rates are bluer.}
    \label{tab:bc-winrate}
    \arrayrulecolor{black}
    \begin{center}
    \begin{tabular}{lcccc|c}
     & POWorkerRush & POLightRush & CoacAI & Mayari & Overall \\
    \arrayrulecolor{black}\specialrule{.5pt}{0pt}{0pt}
    \texttt{basesWorkers8x8A} & \colcell{60} & \colcell{100} & \colcell{90} & \colcell{50} & \colcell{75} \\
    \texttt{FourBasesWorkers8x8} & \colcell{100} & \colcell{100} & \colcell{85} & \colcell{65} & \colcell{88} \\
    \texttt{NoWhereToRun9x8} & \colcell{100} & \colcell{100} & \colcell{83} & \colcell{55} & \colcell{85} \\
    \texttt{basesWorkers16x16A} & \colcell{10} & \colcell{100} & \colcell{100} & \colcell{28} & \colcell{60} \\
    \texttt{TwoBasesBarracks16x16} & \colcell{100} & \colcell{100} & \colcell{43} & \colcell{20} & \colcell{66} \\
    \texttt{DoubleGame24x24} & \colcell{0} & \colcell{100} & \colcell{100} & \colcell{30} & \colcell{58} \\
    \texttt{BWDistantResources32x32} & \colcell{48} & \colcell{100} & \colcell{100} & \colcell{65} & \colcell{78} \\
    \texttt{(4)BloodBath.scmB} & \colcell{100} & \colcell{63} & \colcell{20} & \colcell{40} & \colcell{56} \\
    \arrayrulecolor{black}\specialrule{.5pt}{0pt}{0pt}
    AI Average & \colcell{65} & \colcell{96} & \colcell{78} & \colcell{44} & \colcell{71} \\
    \end{tabular}
    \end{center}
\end{supptable}

\section{Videos of \agentName\ Against Mayari (2021 CoG Winner)}
At \url{https://github.com/sgoodfriend/rl-algo-impls/tree/8230a7c1/papers/cog2024/vsMayariVideos}, we provide videos of
\agentName\ playing against the 2021 CoG competition winner Mayari on each of the Open
maps. \agentName\ is always Player 1 in these videos, thus \agentName's units have a
blue outline while Mayari's units have a red outline. \agentName's units start at the left or
upper part of the map, except in \mapname{(4)BloodBath.scmB} where \agentName\ starts at the
bottom-right. The videos are named after the map they are played on.

\end{document}